\documentclass[10pt,journal,compsoc]{IEEEtran}
\usepackage{amsmath,amsfonts}
\usepackage{algorithmic}
\usepackage{array}
\usepackage[caption=false,font=normalsize,labelfont=sf,textfont=sf]{subfig}
\usepackage{textcomp}
\usepackage{stfloats}
\usepackage{url}
\usepackage{verbatim}
\usepackage{graphicx}
\usepackage{arydshln}
\ifCLASSOPTIONcompsoc
  \usepackage[nocompress]{cite}
\else
  \usepackage{cite}
\fi
\usepackage[colorlinks=true, linkcolor=blue,citecolor=blue,urlcolor=blue]{hyperref}
\usepackage{enumitem}
\usepackage{multirow}
\usepackage{microtype}
\usepackage{multicol}
\usepackage{pifont}
\usepackage[table]{xcolor}
\usepackage{soul}
\usepackage{threeparttable}
\definecolor{lightgray}{RGB}{220,220,220}
\definecolor{lightpurple}{RGB}{230,230,255}
\definecolor{revisionyellow}{RGB}{255,248,195}
\usepackage{booktabs}
\newcommand{\ie}{\textit{i.e.}}
\newcommand{\eg}{\textit{e.g.}}

\newcommand{\ours}{D$^2$MoE }
\newcommand{\pub}[1]{\color{gray}{\tiny{[{#1}]}}}
\newcommand{\best}[1]{\cellcolor{LightCyan2}\textbf{#1}}
\newcommand{\second}[1]{\cellcolor{LavenderBlush1}\underline{#1}}

\def\BibTeX{{\rm B\kern-.05em{\sc i\kern-.025em b}\kern-.08em
    T\kern-.1667em\lower.7ex\hbox{E}\kern-.125emX}}
\usepackage{balance}
\begin{document}
\bstctlcite{BSTcontrol}
\title{Unlocking the Potential of Vision Foundation Models for Multi-Task Dense Prediction via Mixture of Task Adapters}
\author{Yunzhi Zhuge, Xinzhuo Yu, Sitong Gong, Lu Zhang, Pingping Zhang, Huchuan Lu,~\IEEEmembership{Fellow,~IEEE}
        % <-this % stops a space
\IEEEcompsocitemizethanks{\IEEEcompsocthanksitem Yunzhi Zhuge and Xinzhuo Yu contribute equally to this work. 
% This work was supported in part by the National Natural Science Foundation of China under Grants 62406053 and 62441231, in part by the Natural Science Foundation of Liaoning Provincial under Grant 2025-BS-0080, and in part by the 
% the Fundamental Research Funds for the Central Universities under Grant DUT25LAB128.(Yunzhi Zhuge and Xinzhuo Yu contribute equally to this work.)(Corresponding authors: Lu Zhang and Huchuan Lu)
\IEEEcompsocthanksitem  Yunzhi Zhuge, Sitong Gong, Lu Zhang, Pingping Zhang and Huchuan Lu are with the School of Information and Communication Engineering, Dalian University of Technology, Dalian 116081, China. (Email: zgyz@dlut.edu.cn; stgong@mail.dlut.edu.cn; zhangluu@dlut.edu.cn; zhpp@dlut.edu.cn; lhchuan@dlut.edu.cn)
\IEEEcompsocthanksitem Xinzhuo Yu is with the School of Computer Science and Technology, Dalian University of Technology, Dalian 116081, China. (Email: yxz\_3132@mail.dlut.edu.cn)}}
% \protect\\ E-mail: xxx}

\markboth{IEEE Transactions on Pattern Analysis and Machine Intelligence}%
{How to Use the IEEEtran \LaTeX \ Templates}

\IEEEtitleabstractindextext{%
\begin{abstract}
Multi-task dense prediction seeks a single, coherent representation that supports simultaneously discriminative semantic recognition and regularized geometric/structural inference. For vision foundation models, pretrained semantic invariances conflict with the frequency and smoothness biases of dense objectives; joint fine-tuning and single-route adapters therefore compress competing gradients into one update channel, causing negative transfer and unstable cross-task agreement. We frame this as two-level adaptation: task-generic alignment that preserves priors, and task-aware specialization that resolves late-stage conflicts. Based on this principle, we propose \ours, a Mixture of Task Adapters that explicitly factorizes these roles within a frozen DINOv3 backbone. \ours introduces Task-Generic Low-Rank Adaptation (TG-LoRA) in attention projections throughout the network to provide a stable, shared adaptation channel, and a Task-Adaptive Mixture of Experts (TA-MoE) injected only at the final attention block of each stage to enact sparse, task-conditioned perturbations of token mixing in a single forward pass. A Semantic Prior Injector further couples multi-level task features with frozen foundation priors to reinforce semantic coherence while preserving boundary and structural fidelity. Experiments across diverse benchmarks substantiate that this structured factorization yields robust multi-task improvements with a small trainable footprint, establishing a principled and scalable route to prior-preserving, foundation-centric multi-task dense perception.

\end{abstract}

}

\maketitle
\IEEEdisplaynontitleabstractindextext
   
\section{Introduction}
\label{sec:intro}
% 第一段：问题背景：MTDP 进入 Foundation Model 时代
Multi-task dense prediction (MTDP)~\cite{vandenhende2021multi,ye2023taskprompter,jiang2024task} unifies multiple pixel-level tasks (\eg, semantic segmentation~\cite{long2015fully,shelhamer2017fcn,zhao2017pyramid,badrinarayanan2017segnet,chen2018deeplab,zheng2021rethinking,strudel2021segmenter,xie2021segformer,zhou2022rethinking,huang2023criss,zhu2024self}, depth estimation~\cite{liu2015deep,fu2018deep,gordon2019depth,bhat2021adabins,bian2021unsupervised,yang2024depth,yang2024depth,yang2024depthv2}, and saliency detection~\cite{wang2022saliency,liu2016dhsnet,wang2017learning,hou2019deeply,zhuge2019deep,chen2021cnn}) in a single model to share computation, couple semantic/geometric/structural cues, and reduce redundancy relative to task-specific systems. This joint setting further promotes cross-task spatial coherence, especially near boundaries and geometric discontinuities. Recent MTDP increasingly builds on vision foundation models (VFMs)~\cite{caron2021emerging,oquab2023dinov2,simeoni2025dinov3}, using pretrained transformers (e.g., ViT~\cite{ye2022inverted, ye2024invpt++, ye2023taskprompter, ye2023taskexpert, yang2024multi, yangmulti, lu2025swiss} and Swin Transformer~\cite{lin2024mtmamba, lin2025mtmamba++, cao2025enhancing}). As a result, the central problem shifts from learning shared representations to coordinating adaptation of semantically organized foundation features across heterogeneous dense objectives.
Although vision foundation models provide rich, transferable features, MTDP under VFMs faces a core incompatibility: one representation must satisfy objectives with distinct spatial and frequency preferences. This yields three challenges: \textbf{(i) Conflicting Task Priors}. Semantic objectives depend on localized discriminative detail, whereas geometric objectives prefer smooth, globally consistent structure. \textbf{(ii) Prior Erosion}. Naive end-to-end optimization  overwrites transferable invariances, degrading generalization and cross-task consistency. \textbf{(iii) Single-route Adaptation Interference}, \emph{i.e.},forcing all tasks through shared prompts/adapters/decoders projects heterogeneous gradients into one subspace, amplifying negative transfer. These observations motivate conditional, task-aware adaptation that selectively updates the representation to enable task-specific refinement while preserving the shared prior. Selective routing via mixture of experts provides a principled mechanism for allocating distinct update paths to competing objectives and mitigating negative transfer~\cite{fedus2022switch,du2022glam}.

\begin{figure*}
    \centering
    \includegraphics[width=1\linewidth]{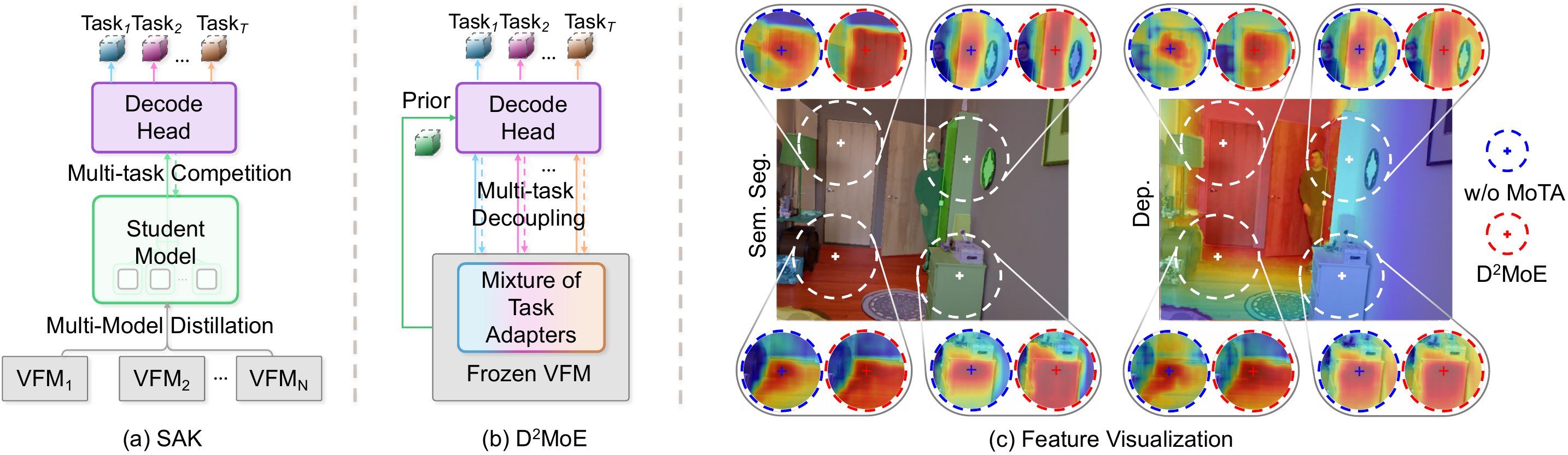}
    \vspace{-5mm}
     % SAK vs. D²MoE.
    \caption{\textbf{Motivation of D$^2$MoE}. (a) SAK distills from multiple VFMs into a shared student, coupling heterogeneous task gradients and incurring multi-teacher overhead. (b) D$^2$MoE instead freezes a single VFM and enforces early task decoupling through a Mixture of Task Adapter (MoTA), allocating sparse, task-conditioned low-rank update routes for specialization without eroding pretrained invariances. (c) Correlation probes (Pearson maps around sampled points) confirm that model with MoTA induces more task-consistent feature organization for both semantic segmentation and depth, improving boundary adherence without sacrificing long-range structural coherence.}
%     (a) SAK relies on multi-model distillation to distill prior knowledge from multiple vision foundation models (VFMs), whereas (b) D²MoE achieves early task decoupling via task-specific adapters on a frozen VFM, delivering higher performance while requiring
% substantially fewer trainable parameters.
% (c) Feature correlation visualization of the Semantic Prior Injector (SPI).
% For semantic segmentation (left) and depth estimation (right), we visualize multi-scale features fused by SPI. At each marked location, we compute the Pearson correlation between the feature at the center point and features within the surrounding circular region. Correlation values are mapped to the jet color map, where warmer colors indicate higher similarity to the center feature and cooler colors indicate lower correlation. Results are compared between w/o MoTA and D$^2$MoE, highlighting improved task-consistent feature structuring.}
    \label{fig:1}
\vspace{-2mm}
\end{figure*}

%第三段：关键转折：论证 MoE 是“逻辑必然解”，而不是可选方案
These observations indicate that once a single shared adaptation path bottlenecks, MTDP needs task-conditioned computation at the adaptation interface while preserving a shared foundation prior rather than hard-partitioning the backbone. \textbf{(i) Unscalable Factorization}. Separate encoders/decoders for each task destroys sharing and is prohibitive for foundation-scale backbones and dense outputs~\cite{ye2022inverted,zhang2025bridgenet}. \textbf{(ii) Delayed Conflict Resolution}. Late-branching heads postpone specialization, leaving upstream features exposed to the same conflicting gradients. \textbf{(iii) Sparse, Task-conditioned Capacity}. Effective MTDP should keep a single backbone but introduce task-dependent specialization via sparsely activated auxiliary parameters. Mixture of experts (MoE) implements this through sparse gating/routing, allocating partially disjoint subspaces per objective without sacrificing the shared representation~\cite{lepikhin2020gshard,riquelme2021scaling,fedus2022switch,du2022glam,zhou2022mixture}.  In contrast, non-MoE MTDP either applies largely uniform encoder adaptation~\cite{agiza2024mtlora,lu2024prompt} or shifts capacity into decoders whose task interaction is weakly coupled to the backbone features being adapted~\cite{lin2024mtmamba,lin2025mtmamba++}.

\begin{figure}
    \centering
    \includegraphics[width=1\linewidth]{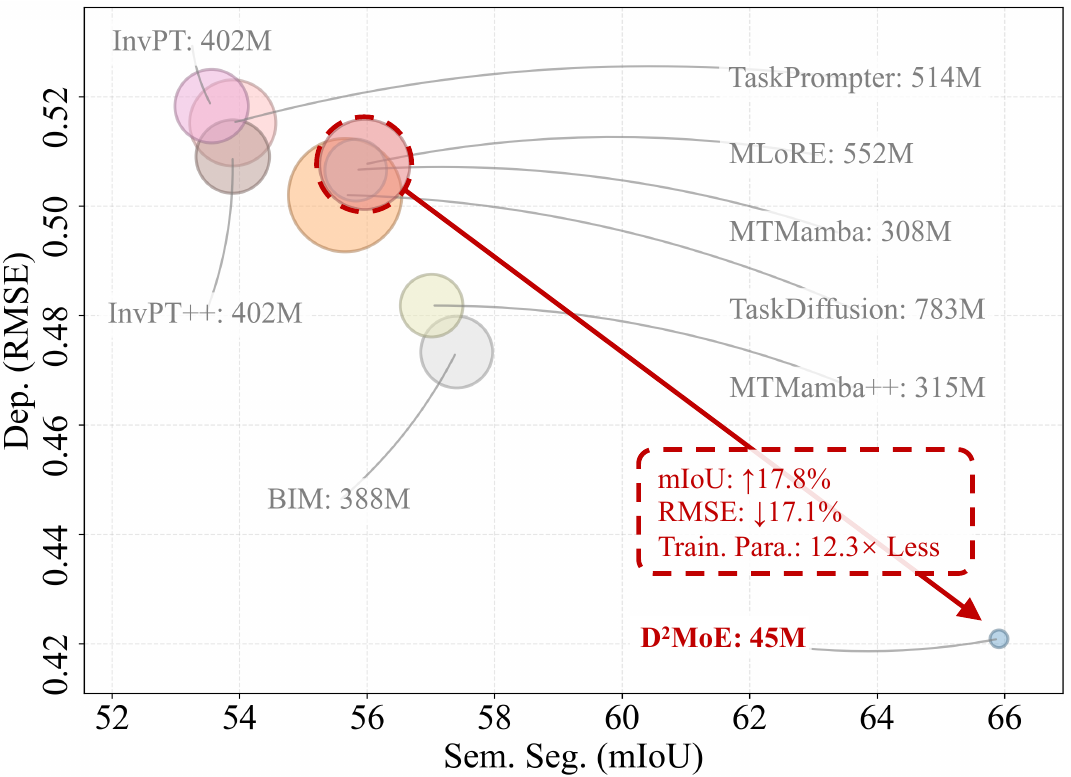}
    % \put(-150,85){\scriptsize \cite{zhang2025bridgenet}}
    \put(-182,108){\scriptsize \cite{ye2024invpt++}}     %InvPT++
    \put(-189.5,172.5){\scriptsize \cite{ye2022inverted}}    %InvPT
    \put(-157,64){\scriptsize \cite{cao2025enhancing}}      %BIM
    \put(-22,164){\scriptsize \cite{ye2023taskprompter}}  %TaskPrompter
    % \put(-26,157){\scriptsize \cite{ye2023taskexpert}}
    \put(-38,146.2){\scriptsize \cite{yang2024multi}}  %MLoRE
    \put(-29,127.6){\scriptsize \cite{lin2024mtmamba}}   %MTMamba
    \put(-21,109){\scriptsize \cite{yangmulti}}  %TaskDiffusion
    \put(-21,90.8){\scriptsize \cite{lin2025mtmamba++}} %MTMamba++
	% \put(-443.5,219){\scriptsize \S\ref{sec:2}}
    \caption{\textbf{Performance vs. Model Complexity}. Comparison of \ours against state-of-the-art methods on  NYUD-v2~\cite{silberman2012indoor}. \ours achieves superior performance (higher mIoU and lower RMSE) despite a substantial reduction in the total number of trainable parameters. The trainable-parameter count varies with the task configuration, with 45M on NYUD-v2 (shown here) and 54M on Pascal-Context (as shown in Tab.~\ref{Pascal Context Comparison}.)}
    \label{fig:2}
\end{figure}

% \textcolor{blue}{In parallel, parameter-efficient multi-task adaptation has recently attracted increasing attention. Methods such as MTLoRA, Polyhistor, VMT-Adapter, TADFormer, and DiTASK improve multi-task transfer by introducing task-specific low-rank updates, hypernetwork-based sharing, once-for-all adapters, dynamic task filters, or singular-value transformations. These studies demonstrate that task-aware PEFT is effective for multi-task vision models. However, they mainly focus on how to parameterize or generate task-specific updates, while the adaptation of frozen vision foundation models for dense prediction raises a different question: where and how should task-conditioned specialization be introduced so that pretrained semantic priors are preserved rather than overwritten?}
Several recent parameter-efficient and conditional multi-task adaptation methods are closely related to this direction. 
Polyhistor~\cite{liu2022polyhistor} and VMT-Adapter~\cite{xin2024vmt} improve multi-task transfer by sharing lightweight adaptation modules while preserving task-specific knowledge, and TADFormer~\cite{baek2025tadformer} further introduces task-adaptive dynamic filtering for input-conditioned task adaptation. 
DiTask~\cite{mantri2025ditask} studies multi-task fine-tuning from a weight-geometry perspective by preserving pretrained singular-vector structures and modulating singular values. 
These methods validate the importance of parameter-efficient multi-task adaptation, yet most of them focus on designing general adaptation modules or task-conditioned filters, rather than explicitly locating where task conflicts should be resolved inside a frozen vision foundation model for dense prediction. 
In parallel, MoE-based MTL methods such as M$^3$ViT~\cite{fan2022m3vit}, Mod-Squad~\cite{chen2023mod}, AdaMV-MoE~\cite{chen2023adamv}, TaskExpert~\cite{ye2023taskexpert} and MLoRE~\cite{yang2024multi} introduce conditional expert selection to improve cooperation-specialization trade-offs. 
However, existing MoE designs typically insert trainable experts throughout the backbone, assemble task features in the decoder, or target compact task-specific execution, which may either perturb pretrained representations or leave task conflicts insufficiently addressed.
% Our D$^2$MoE differs by formulating MTDP with VFMs as a prior-preserving two-level adaptation problem: TG-LoRA performs task-generic alignment across the frozen backbone, whereas TA-MoE introduces sparse task-conditioned low-rank perturbations only at the final attention block of each stage, directly specializing the token-mixing operator while retaining a single frozen foundation prior.}

%第四段：我们的核心洞察：任务能力应分层、分角色组织
Motivated by the above analysis, we view the MTDP bottleneck under frozen VFMs as requiring two complementary capabilities: \textbf{task-generic adaptation} and \textbf{task-aware specialization}. Task-generic adaptation aligns the foundation representation to the downstream data distribution while maintaining transferable semantic prior, whereas task-aware specialization explicitly resolves conflicts induced by heterogeneous dense objectives. 
% In SAK~\cite{lu2025swiss}, a shared teacher-agnostic stem with teacher-specific adapter paths distills complementary biases from multiple VFMs and fuses them via task routers for multi-task prediction. 
A closely related foundation-model-driven approach is SAK~\cite{lu2025swiss}, which uses a shared teacher-agnostic stem with teacher-specific adapter paths to distill complementary biases from multiple VFMs and fuses them via task routers for multi-task prediction. However, the student is optimized to satisfy heterogeneous teacher feature spaces, yielding a compromise representation that weakens domain adaptation, while specialization remains teacher-conditioned rather than explicitly resolving task-level gradient conflicts. Practically, requiring $\geq$3 VFM teachers incurs $O(K)$ compute/memory (multiple frozen backbones, matching losses) during training, making it bulky for deployment (as shown in Fig.~\ref{fig:1}(a)).

To address the above limitations, we propose \textbf{D$^2$MoE}, a foundation model driven framework for multi-task dense prediction. 
Unlike prior MoE-based MTDP methods that mainly emphasize task routing and expert specialization, \ours uses MoE as part of a lightweight adapter design to introduce task-adaptive updates without tuning the VFM backbone. % R2_C1
D$^2$MoE builds on a single frozen VFM backbone and performs multi-task decoupling through a unified Mixture of Task Adapters (MoTA) design (as shown in Fig.~\ref{fig:1}(b)), which combines Task-Generic Low-rank Adaptation (TG-LoRA) with Task-Adaptive Mixture of Experts (TA-MoE). TG-LoRA is applied across backbone layers to achieve stable and parameter-efficient adaptation, while TA-MoE is selectively placed at the final layer of each stage to enable sparse, task-aware specialization within a single forward pass. In addition, D$^2$MoE incorporates semantic priors from DINOv3 to guide cross-layer feature integration, facilitating both global context modeling and fine-grained detail preservation. As a result, D$^2$MoE achieves state-of-the-art performance on standard multi-task dense prediction benchmarks with minimal additional parameters (as shown in Fig.~\ref{fig:2}).
% D$^2$MoE introduces a unified Mixture of Task Adapters (MoTA) design that combines Task-Generic Low-rank Adaptation (TG-LoRA) with Task-Adaptive Mixture of Experts (TA-MoE). TG-LoRA is applied across backbone layers to achieve stable and parameter-efficient adaptation, while TA-MoE is selectively placed at the final layer of each stage to enable sparse, task-aware specialization within a single forward pass. In addition, D$^2$MoE incorporates semantic priors from DINOv3 to guide cross-layer feature integration, facilitating both global context modeling and fine-grained detail preservation. As a result, D$^2$MoE achieves state-of-the-art performance on standard multi-task dense prediction benchmarks with minimal additional parameters.

%Contributions
Our contributions are summarized as follows.
\begin{itemize}[leftmargin=1.8em,labelsep=0.6em]
\item[1)] We formalize multi-task dense prediction with frozen vision foundation models as a two-level adaptation problem, disentangling task-generic alignment from task-aware specialization. This perspective explains negative transfer under single-route adaptation and motivates sparse, task-conditioned computation.
\item[2)] We propose \ours, a Mixture of Task Adapters architecture that combines Task-Generic Low-Rank Adaptation with Task-Adaptive Mixture of Experts. The design introduces sparse, late-stage task specialization while preserving  semantic priors within a single backbone.
% DINOv3-based   pretrained
\item[3)] Extensive experiments on PASCAL-Context, NYUD-v2, and Cityscapes show that \ours achieves state-of-the-art multi-task performance, delivering the highest aggregated gains across benchmarks while requiring substantially fewer trainable parameters.
% \item[1)] We present D$^2$MoE, the first framework that systematically integrates DINOv3 with task-adaptive MoE for multi-task dense prediction.
% \item[2)] We propose a layered MoE design that explicitly separates task-generic adaptation from task-specific specialization, providing a principled solution to MTDP in the foundation model regime.
% \item[3)] Extensive experiments demonstrate that D$^2$MoE achieves state-of-the-art performance across multiple benchmarks while requiring less than 20\% additional trainable parameters.
\end{itemize}

The remainder of this paper is structured as follows. Sec.~\ref{sec:2} surveys related works on multi-task dense prediction, vision foundation models, and mixture of experts. Sec.~\ref{sec:3} introduces \ours, including the Mixture of Task Adapters and the Semantic Prior Injector for prior-guided decoding. Sec.~\ref{sec:4} presents the experimental protocol and results on PASCAL-Context, NYUD-v2, and Cityscape.
% , with state-of-the-art comparisons, ablations, and diagnostic analyses. 
Sec.~\ref{sec:5} concludes with key findings and potential implications.

\section{Related Works}
\label{sec:2}

\subsection{Multi-task Dense Prediction}
Multi-task dense prediction, which jointly addresses semantic segmentation, depth estimation, and auxiliary tasks for scene understanding, has substantively progressed via relational modeling, efficient architectures, and generative formulations~\cite{misra2016crossstitch,kokkinos2017ubernet,xu2018pad,liu2019end,vandenhende2020mti,zamir2018taskonomy,vandenhende2021multi,xu2022mtformer,yang2023contrastive,ye2023taskprompter,lu2024prompt}. ATRC~\cite{bruggemann2021exploring} explicitly formalizes inter-task dependencies to improve transferable generalization. MTPSL~\cite{li2022learning} tackles incomplete annotations in a partially-supervised setting, enabling resilient learning under  partial labels. Transformer-based designs such as the InvPT enable hierarchical feature fusion and granular scene parsing \cite{ye2022inverted,ye2023taskprompter}.
Efficiency has been advanced by cross-task distillation, epitomized by KDAM \cite{xu2023multi}. MLoRA employs a mixture of low-rank experts for parameter-efficient, per-task routing that scales to heterogeneous dense outputs \cite{yang2024multi}. Synergy- and compactness-oriented approaches include SEM, which promotes salutary task interactions beyond naive sharing \cite{huang2024going}, and Bi-MTPD, which binarizes predictors for compact models with negligible performance loss \cite{shang2024efficient}. State-space innovations such as MTMamba~\cite{lin2024mtmamba} capture long-range dependencies for enhanced multi-task understanding, whereas MTMamba++~\cite{lin2025mtmamba++} streamlines decoding to improve efficiency. BridgeNet~\cite{zhang2025bridgenet} introduces bridge embeddings to facilitate comprehensive cross-task feature interactions. Collectively, these lines of work converge on on holistic, adaptive and resource-conscious frameworks. Collectively, these lines of work converge holistic and resource-conscious frameworks.

\subsection{Parameter Efficient Multi-task Adaption}
Parameter-efficient multi-task adaptation studies how to transfer pretrained models to multiple downstream tasks with limited trainable parameters, making it complementary to MTDP-specific architectures. 
Polyhistor~\cite{liu2022polyhistor} uses decomposed hypernetworks and layer-wise scaling kernels to share knowledge without full task-wise fine-tuning. 
VMT-Adapter~\cite{xin2024vmt} introduces a once-for-all adapter that combines shared cross-task knowledge with lightweight task-specific extraction, yielding nearly task-number-independent cost. TADFormer~\cite{baek2025tadformer} performs task-aware adaptation through parameter-efficient prompting and dynamic task filters conditioned on task contexts. 
DiTask~\cite{mantri2025ditask} preserves pretrained weight geometry via diffeomorphic transformations of singular values for shared and task-specific modulation. 
These methods show that separating shared adaptation from task-specific modulation is effective for multi-task transfer. 
Different from them, \ours targets frozen VFMs for MTDP by coupling global TG-LoRA alignment with sparse TA-MoE specialization at late-stage attention projections, where heterogeneous dense objectives most strongly diverge.

\subsection{Vision Foundation Models}
Vision foundation models (VFMs) have revolutionized computer vision by learning generalizable representations from vast unlabeled datasets, enabling transfer to diverse downstream tasks~\cite{awais2025foundation}. Pioneering architectures like Vision Transformers (ViT)~\cite{dosovitskiy2020image} scaled transformers for image processing, achieving state-of-the-art performance on benchmarks such as ImageNet. Self-supervised approaches, including Masked Autoencoders (MAE)~\cite{he2022masked}, reconstruct masked image patches to capture semantic features without annotations. DINO~\cite{caron2021emerging} employs self-distillation to emerge invariant representations, fostering emergent properties like object segmentation. Multimodal models such as CLIP align vision and language embeddings for zero-shot classification and retrieval. Subsequent iterations like DINOv2~\cite{oquab2023dinov2} refine training with curated data and improved objectives, yielding robust features across scales. The Segment Anything Model (SAM)~\cite{kirillov2023segment} introduces promptable segmentation for arbitrary objects, advancing interactive vision applications. Most recently, DINOv3~\cite{simeoni2025dinov3} extends these capabilities with enhanced multi-task potential through advanced self-supervision techniques. Accordingly, this work targets prior-preserving, parameter-efficient task-conditioned adaptation of VFMs for multi-task dense prediction.
% While these models excel in single-task fine-tuning, efficient multi-task adaptation remains a key challenge.

\subsection{Mixture of Experts}
Mixture of experts (MoE)  has emerged as a $\textit{de\ \ facto}$ paradigm for conditional computation: a router sparsely activates a small subset of experts per token/sample, increasing capacity without proportional FLOPs. Foundational designs such as Switch Transformer~\cite{fedus2022switch} and GLaM~\cite{du2022glam} establish scalable sparse routing, while Expert-Choice routing~\cite{zhou2022mixture} and DeepSpeed-MoE~\cite{rajbhandari2022deepspeed} show that routing policies and systems support are central to utilization and efficiency. 
% MoE’s key leverage is sparse, well-balanced specialization under practical training/inference constraints, which our method inherits in the adapter space.
% Mixture of experts (MoE) has emerged as a $\textit{de\ \ facto}$ paradigm for conditional computation: a router dynamically selects a sparse subset of experts per token/sample, thereby scaling model capacity without a commensurate increase in FLOPs. Switch Transformer~\cite{fedus2022switch} simplifies sparse routing to a top-1 design while addressing training stability, and GLaM~\cite{du2022glam} demonstrates that sparsely activated MoE can scale language modeling efficiently at scale. Subsequent work highlights that routing itself is a key design axis: Expert-Choice routing~\cite{zhou2022mixture} redefines assignment by letting experts select tokens, improving utilization and load balancing. From system perspective, DeepSpeed-MoE~\cite{rajbhandari2022deepspeed} makes expert parallelism practical and improves end-to-end training/inference efficiency for large sparse models.

In vision and multimodal foundation models, MoE is used to reduce domain/task interference while retaining scalable capacity. V-MoE~\cite{riquelme2021scaling} adapts sparse experts to ViTs, LIMoE~\cite{mustafa2022multimodal} shows modality-aware partitioning under contrastive learning, and Uni-Perceiver-MoE~\cite{zhu2022uni} integrates conditional experts into unified generalist architectures. More recent multimodal variants route over heterogeneous expert pools for generalist MLLMs or vision experts (MoME~\cite{shen2024mome}, MoVA~\cite{zong2024mova}), with Uni-MoE~\cite{li2025uni} further systematizing scaling in unified multimodal LLMs.
% In vision and multimodal foundation models, MoE is increasingly used to mitigate interference across domains while retaining scalable capacity. V-MoE~\cite{riquelme2021scaling} extends sparse experts to Vision Transformers and shows competitive scaling for image recognition, while LIMoE~\cite{mustafa2022multimodal} demonstrates that experts can naturally partition modalities under multimodal contrastive learning. For generalist architectures, Uni-Perceiver-MoE~\cite{zhu2022uni} integrates conditional experts into a unified multi-task/multi-modal model to reduce cross-task conflicts. Recent multimodal MoE variants explicitly route among heterogeneous expert pools, \eg, MoME~\cite{shen2024mome} for generalist MLLMs and MoVA~\cite{zong2024mova} for instruction-conditioned selection/fusion of vision experts. Uni-MoE~\cite{li2025uni} further systematizes MoE scaling for unified multimodal LLMs. 

MoE is also effective for multi-task learning and parameter-efficient specialization, where conditional routing allocates capacity to tasks with higher complexity or conflict. Vision MTL approaches route along task-conditioned expert pathways with varied sharing/gating (M$^3$ViT~\cite{fan2022m3vit}, Mod-Squad~\cite{chen2023mod}, AdaMV-MoE~\cite{chen2023adamv}, TaskExpert~\cite{ye2023taskexpert}), and MLoRE~\cite{yang2024multi} scales experts for dense prediction via low-rank parameterization. Recent work reframes MoE as adapter-style tuning: MoE-Adapters~\cite{yu2024boosting} sparsely selects from a fixed adapter pool (with OOD fallback), and MoE-Adapters++~\cite{yu2025moe_adapter++} introduces dynamic adapters and learned expert selection. Inspired by adapter-style MoE, we route among lightweight task adapters atop frozen DINOv3 features, concentrating specialization in small modules while preserving backbone transferability.
\section{Methodology}
\label{sec:3}

\begin{figure*}
    \centering
    \includegraphics[width=1\linewidth]{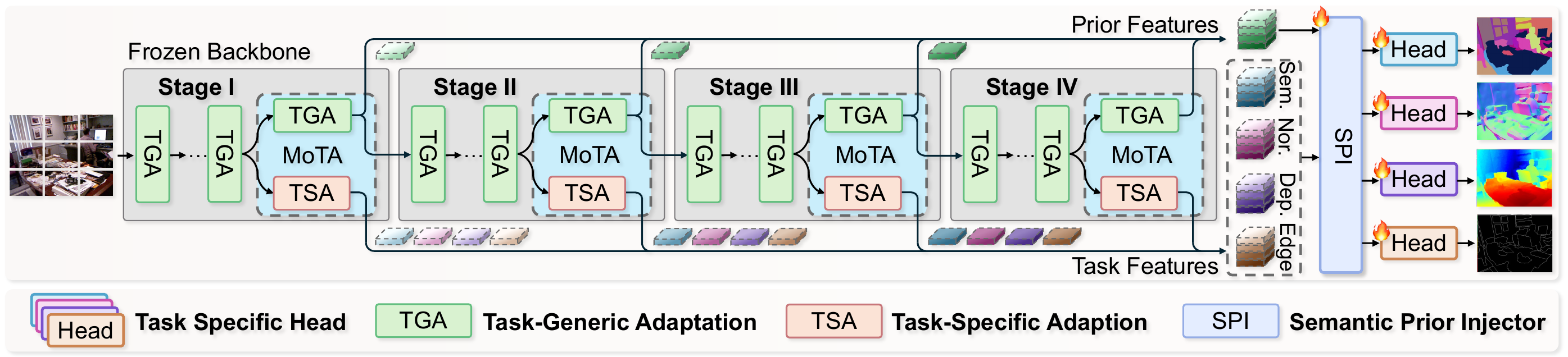}
    \caption{\textbf{Overall architecture of D$^2$MoE}.(\S \ref{sec:3.1}) Built upon a frozen DINOv3 backbone, \ours organizes MoTA across four stages to preserve shared priors through task-generic adaptation and introduce localized task-specific adaption at stage endings. SPI fuses prior and task features, followed by lightweight heads for multi-task dense predictions.}
    % Built upon a frozen DINOv3, \ours introduces a structured Mixture of Task Adapters (MoTA) and a Semantic Prior Injector (SPI).
    % Specifically, }
    % Task-Generic LoRA (TG-LoRA) is inserted into all ViT blocks to robustly align pretrained representations with downstream data distributions while preserving semantic invariances. In complement, Task-Adaptive MoE (TA-MoE) is placed at the final attention layer of each stage to resolve late-stage conflicts through sparse, task-conditioned expert routing within a single forward.}
    % D$^2$MoE is built upon the DINOv3 backbone and consists of two key modules: the Mixture of Task Adapters (MoTAs) and the Semantic Prior Injector. To be more specific, Task-Generic LoRA (TG-LoRA) is inserted into all ViT blocks to achieve xxx, while Task-Adaptive MoE (TA-MoE) is placed at the final layer of each stage with a TG-LoRA to compose MoTA.}
    \label{fig:pipeline}
\end{figure*}

\begin{figure*}
    \centering
    \includegraphics[width=1\linewidth]{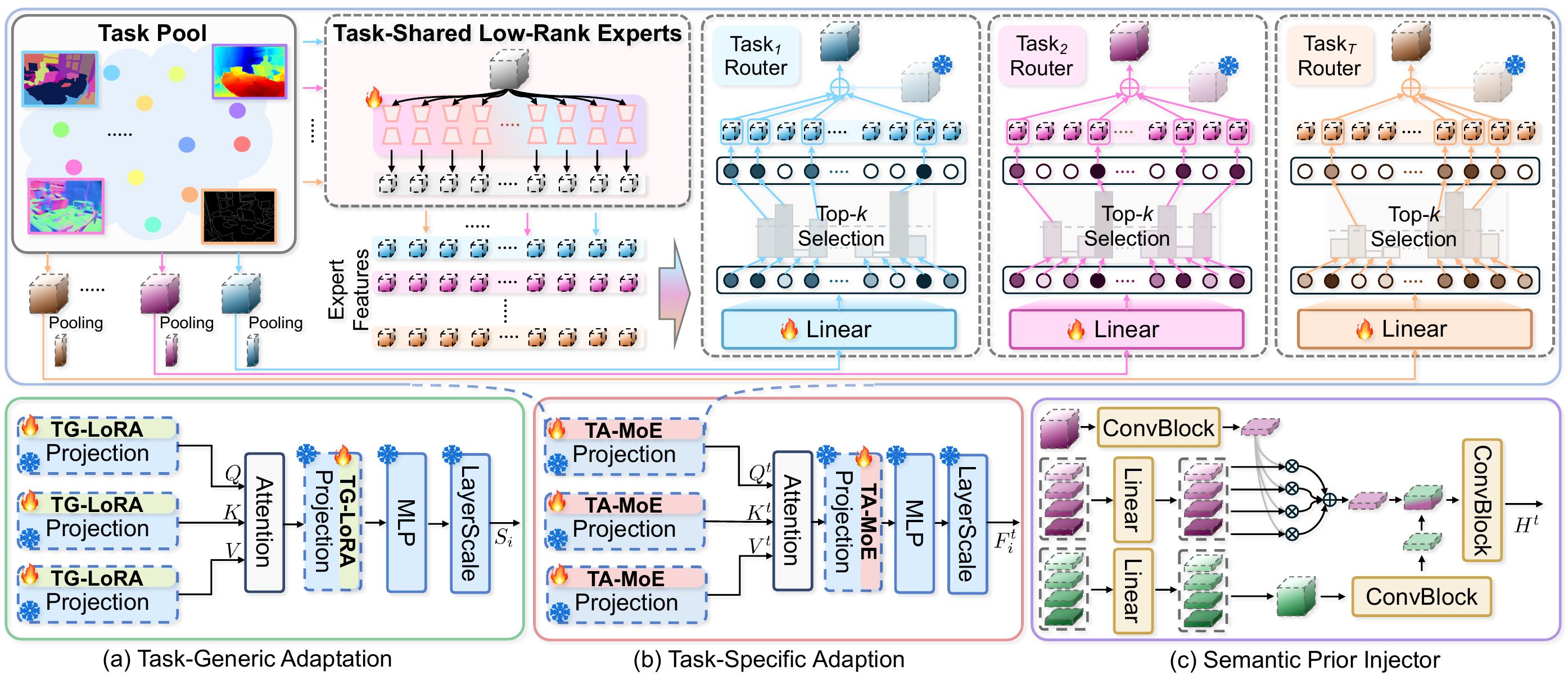}
    \caption{\textbf{Illustration of the key components in D²MoE}. (\S \ref{sec:3.2} and \ref{sec:3.3}) (a) Task-Generic Adaptation inserts TG-LoRA into attention projections to provide shared, low-rank adaptation while keeping the DINOv3 backbone frozen. (b) Task-Specific Adaptation employs TA-MoE, where task routers select Top-k low-rank experts to produce sparse, task-conditioned updates for attention projections. (c) The Semantic Prior Injector fuses multi-level task features with frozen semantic priors to enhance semantic consistency and structural coherence for dense prediction tasks.}
    \label{fig:modules}
\end{figure*}

% This section introduces D$^2$MoE, a parameter-efficient multi-task dense prediction framework that freezes DINOv3 to preserve its pretrained representation and trains only lightweight task-aware modules. Sec.~\ref{sec:3.1} gives an overview of the pipeline. Sec.~\ref{sec:3.2} describes our Mixture of Task Adapters, combining Task-Generic Low-Rank Adaption (inserted into Q/K/V and output projections) with Task-Adaptive Mixture of Experts activated only at the final attention layer of each stage, where a gate selects Top-$k$ low-rank experts for task-specific feature modulation. Sec.~\ref{sec:3.3} presents the Semantic Prior Injector (SPI), which fuses multi-level task features with semantic priors to enhance semantic consistency and structure coherence across tasks.
{
\looseness=-1
This section presents \ours, a parameter-efficient framework for adapting frozen vision foundation models to multi-task dense prediction. Sec.~\ref{sec:3.1} first provides an overview of the pipeline, including the frozen DINOv3 backbone, the Mixture of Task Adapters (MoTA), and the Semantic Prior Injector (SPI). Sec.~\ref{sec:3.2} then details MoTA, which decouples shared adaptation from task-conditioned specialization. Finally, Sec.~\ref{sec:3.3} introduces SPI for prior-guided multi-scale decoding.
\par
}

\subsection{Framework Overview}
\label{sec:3.1}
Fig.~\ref{fig:pipeline} illustrates the overall architecture of \ours. Given an input image \(I \in \mathbb{R}^{H \times W \times 3}\), we employ a frozen DINOv3 backbone as the shared feature extractor. The backbone provides two types of representations: task-agnostic semantic priors \(\{S_i\}_{i=1}^{4}\), which retain the pretrained foundation knowledge, and task-adapted features \(\{F_i^t\}_{i=1}^{4}\) for each downstream task \(t \in \{1,\ldots,T\}\).

To adapt the frozen backbone without full fine-tuning, we insert the proposed Mixture of Task Adapters (MoTA) into the attention blocks. MoTA serves as the only backbone-side trainable adaptation path and produces task-aware multi-level features while preserving the shared VFM prior. The adapted features are then passed to the Semantic Prior Injector (SPI), which fuses multi-level task features with frozen semantic priors for prior-guided decoding. Finally, lightweight task-specific heads map the refined representations to the corresponding prediction spaces, yielding task-wise outputs \(\{O_t\}_{t=1}^{T}\).

\subsection{Mixture of Task Adapters}
\label{sec:3.2}
MoTA is designed to address the single-route adaptation bottleneck in frozen VFM-based multi-task dense prediction. Instead of forcing all tasks to share an identical adaptation subspace, MoTA factorizes backbone adaptation into two complementary components: a shared low-rank pathway for stable task-generic alignment, and a sparse task-conditioned pathway for resolving task-specific conflicts. This design allows \ours to retain the semantic priors of the frozen DINOv3 while introducing lightweight conditional capacity only where task divergence needs to be modeled. The following sections describe the two components in detail.

\subsubsection{Task-Generic Low-Rank Adaptation}
\label{sec:3.2.1}
To obtain parameter-efficient transfer while preserving the semantic geometry encoded by the DINOv3 backbone, we introduce \emph{Task-Generic Low-Rank Adaptation} (TG-LoRA) inside the self-attention modules. In contrast to full fine-tuning, TG-LoRA keeps all pretrained backbone weights frozen and learns only structured low-rank increments on a small set of high-impact linear maps. This constrained re-parameterization provides a stable, task-agnostic adaptation channel that aligns the foundation representation to downstream data distributions with reduced risk of representation drift and cross-task interference.

TG-LoRA is inserted exclusively into the query, key, value projections and the attention output projection of each Transformer block. Let $X\in\mathbb{R}^{N\times C}$ denote the input token matrix ($N$ tokens, embedding dimension $C$). Under a pre-normalization design, we first compute:
\begin{equation}
\tilde{X}=\mathrm{LN}(X).
\end{equation}
For $P\in\{Q,K,V\}$, we express the projection as a frozen base weight plus a rank-$r$ residual update:
\begin{equation}
    \Delta W_{P}=B_{P}A_{P},
\end{equation}
\begin{equation}
P=\tilde{X}\bigl(W_{P}+s_{P}\Delta W_{P}\bigr),
\label{eq:tg_lora_qkv}
\end{equation}
where $W_{P}\in\mathbb{R}^{C\times C}$ is frozen, while $A_{P}\in\mathbb{R}^{C\times r}$ and $B_{P}\in\mathbb{R}^{r\times C}$ are learnable with $r\ll C$. The scalar $s_{P}$ controls the effective update magnitude; a common choice is $s_{P}=\alpha_{P}/r$ to keep the perturbation scale approximately stable as $r$ varies. This can be interpreted as learning a constrained perturbation of the form $W_{P}\mapsto W_{P}+s_{P}B_{P}A_{P}$ without modifying the pretrained weights.

The projected features are then processed by multi-head self-attention. Let $h$ denote the number of heads and $d_h=C/h$ the per-head dimension. After reshaping into head-wise tensors $\{Q^{(m)},K^{(m)},V^{(m)}\}_{m=1}^{h}$ with $Q^{(m)},K^{(m)},V^{(m)}\in\mathbb{R}^{N\times d_h}$, each head computes:
\begin{equation}
H^{(m)}=\mathrm{Softmax}\!\left(\frac{Q^{(m)}(K^{(m)})^{\top}}{\sqrt{d_h}}\right)V^{(m)},
\qquad m=1,\dots,h,
\label{eq:mha}
\end{equation}
and the heads are concatenated to form $Z=\mathrm{Concat}(H^{(1)},\dots,H^{(h)})\in\mathbb{R}^{N\times d}$.

TG-LoRA is further applied to the attention output projection that maps contextualized tokens back into the shared embedding space:
\begin{equation}
Z' = Z\bigl(W_{O}+s_{O}B_{O}A_{O}\bigr),
\label{eq:tg_lora_out}
\end{equation}
where $W_{O}\in\mathbb{R}^{d\times d}$ is frozen and $(A_{O},B_{O})$ are learnable rank-$r$ adapters. The attention sub-layer output is then obtained through the residual pathway:
\begin{equation}
Y = X + Z'.
\end{equation}

The FFN and LayerScale follow the original DINOv3 design and are kept frozenThe TG-LoRA augmented output is then propagated to the next Transformer block. To preserve the pretrained VFM's general-purpose representation, semantic priors $\{S_i\}_{i=1}^{4}$ are extracted from the corresponding unmodified DINOv3 blocks applied to the same input $X$, ensuring that the priors reflect task-agnostic backbone features rather than adaptation-induced artifacts.
% ; we omit their standard forms for brevity. 
% The output of each TG-LoRA–augmented block is propagated to the next Trasformer block. In contrast, the semantic prior features $\{S_i\}_{i=1}^{4}$ are extracted from the input $X$ processed by the corresponding original DINOv3 Transformer blocks without TG-LoRA, so as to fully preserve the robust and task-agnostic feature perception of the pretrained VFM.
% The subsequent feed-forward network (FFN) and LayerScale block follow the original DINOv3 formulation and remain frozen. We omit their explicit expressions for brevity. We take the output of the final attention block at each stage as the semantic prior feature, yielding $\{S_i\}_{i=1}^{4}$.

By restricting TG-LoRA to $\{W_Q,W_K,W_V,W_O\}$, the trainable degrees of freedom primarily modulate (i) the token-to-token interaction geometry via $(Q,K)$ and (ii) the injection of contextual evidence back into the embedding space via $(V,W_O)$, while leaving the typically more representationally disruptive FFN untouched. As a result, TG-LoRA acts as a globally shared, task-generic adaptation mechanism that improves robustness to domain shifts while preserving the semantic organization of the foundation model, providing a stable substrate for subsequent task-adaptive specialization.

\subsubsection{Task-Adaptive Mixture of Experts}
\label{sec:3.2.2}
TG-LoRA provides a shared, task-agnostic adaptation route that preserves the backbone’s semantic prior. Nevertheless, heterogeneous dense prediction objectives often demand incompatible token-interaction patterns at late representation levels, where high-level semantics are consolidated and task divergence becomes most pronounced. To explicitly factorize these late-stage conflicts without fragmenting the backbone, we introduce \emph{Task-Adaptive Mixture of Experts} (TA-MoE) as a conditional specialization mechanism.

In contrast to MoE designs that distribute experts broadly across backbone layers or postpone task differentiation to decoder heads~\cite{chen2023mod,fan2022m3vit}, TA-MoE is injected only into the final self-attention block of each DINOv3 stage and acts directly on the attention linear maps (Q/K/V projections and the output projection). This placement targets the token-mixing operator where task-dependent contextual aggregation is most sensitive, while keeping earlier layers fully shared for stable representation learning.

Let $H^{s}\in\mathbb{R}^{N\times C}$ denote the token features entering the last attention block of stage $s$, where $N$ is the number of tokens and $C$ is the channel dimension. For a $T$-task setting, we introduce task-indexed inputs $\{H^{s,t}\}_{t=1}^{T}$ as notational replicas initialized by broadcasting $H^s$, so that task-conditioned operators can be written uniformly, while the shared backbone features $H^s$ are computed only once.

A task-conditioned router predicts a sparse gating distribution over $E$ candidate experts from a global stage summary. We first compute a compact descriptor via token-wise pooling:
\begin{equation}
\bar{h}^{\,t}=\mathrm{Pool}\!\left(H^{s,t}\right)\in\mathbb{R}^{C},
\end{equation}
and obtain expert logits using a lightweight task-specific linear map:
\begin{equation}
a^{t}=R^{t}\bar{h}^{\,t}\in\mathbb{R}^{E},
\end{equation}
where $R^{t}\in\mathbb{R}^{E\times C}$. Sparse activation is enforced by selecting the Top-$k$ experts ($k\ll E$) and normalizing on the active set:
\begin{equation}
\pi^{t}=\mathrm{Softmax}\!\bigl(\mathrm{Top}\text{-}k(a^{t})\bigr)\in\Delta^{E},
\end{equation}
where $\Delta^{E}$ denotes the probability simplex with zeros on inactive experts.

Each expert is implemented as a low-rank residual update. For each attention projection $P\in\{Q,K,V,O\}$, the $e$-th expert is parameterized by factors
$A^{(e)}_{P}\in\mathbb{R}^{C\times r}$ and $B^{(e)}_{P}\in\mathbb{R}^{r\times C}$ with rank $r\ll C$. The task-specific residual to the frozen projection $W_{P}\in\mathbb{R}^{C\times C}$ is:
\begin{equation}
\Delta W^{t}_{P}=\frac{\alpha}{r}\sum_{e=1}^{E}\pi^{t}_{e}\,B^{(e)}_{P}A^{(e)}_{P},
\end{equation}
with scaling factor $\alpha$. Accordingly, task-specialized projections are computed as:
\begin{equation}
P^{t}=H^{s,t}\bigl(W_{P}+\Delta W^{t}_{P}\bigr),\qquad P\in\{Q,K,V\}.
\end{equation}

The task-adaptive $Q^{t}$, $K^{t}$, and $V^{t}$ are then processed by standard multi-head self-attention:
\begin{equation}
Z^{t}=\mathrm{MSA}(Q^{t},K^{t},V^{t})
=\mathrm{Softmax}\!\left(\frac{Q^{t}{K^{t}}^{\top}}{\sqrt{d_h}}\right)V^{t},
\end{equation}
where $d_h$ denotes the per-head dimension (head splitting/concatenation is omitted for brevity). TA-MoE is further applied to the attention output projection to complete a coherent task-conditioned attention pathway:
\begin{equation}
\widetilde{H}^{s,t}=Z^{t}\bigl(W_{O}+\Delta W^{t}_{O}\bigr),
\end{equation}
where $W_{O}$ is the frozen output projection and $\Delta W^{t}_{O}$ is the routed low-rank residual defined above.

The resulting tokens are processed by the standard residual connections, feed-forward transformation, and LayerScale of DINOv3, yielding the task-specific stage-wise outputs $\{F^{t}_{i}\}_{i=1}^{4}$. Overall, TA-MoE interprets task divergence as a conditional perturbation of the token-mixing operator: the backbone projections remain frozen to preserve a shared semantic prior, while a sparse mixture of low-rank experts injects task-selective capacity precisely at the late layers where conflicts are most acute. By restricting conditional computation to the final attention block of each stage, we obtain a clean separation between globally shared representation learning and task-specific refinement, achieving effective disentanglement with minimal additional overhead.

\subsection{Semantic-Prior Guided Decoding}
\label{sec:3.3}

TG-LoRA and TA-MoE adapt the shared backbone for multi-task dense prediction, yet the final outputs still depend on a decoder that maps adapted features to task-specific label spaces.
Most VFM-style decoders remain shallow and task-agnostic, typically performing FPN-like fusion without explicitly exploiting the high-level semantic structure already encoded in foundation models.
We thus propose a semantic-prior guided decoder that integrates task-adapted features with frozen DINOv3 representations, enabling prior-aware multi-scale aggregation and task-conditioned refinement.

\subsubsection{Semantic Prior Injection}

For each task $t$, let $\{X^{t}_{i}\}_{i=1}^{4}$ be the task-adapted features from four backbone stages,
where $X^{t}_{i}\in\mathbb{R}^{C\times H\times W}$,
and let $\{S_{i}\}_{i=1}^{4}$ be the aligned semantic priors extracted from DINOv3,
with $S_{i}\in\mathbb{R}^{C\times H\times W}$.
The Semantic Prior Injector (SPI) performs channel alignment, task-conditioned multi-scale fusion, and semantic prior injection.

Since features from all stages share the same spatial resolution and channel dimensions, no explicit spatial rescaling is required.
We project both task features and semantic priors $C$ using independent $1\times1$ convolutions:
\begin{equation}
\tilde{X}^{t}_{i} = \phi_{x}(X^{t}_{i}),
\end{equation}
\begin{equation}
\tilde{S}_{i} = \phi_{s}(S_{i}),
\end{equation}
where $\tilde{X}^{t}_{i}, \tilde{S}_{i} \in \mathbb{R}^{C\times H\times W}$. To aggregate task features across scales, we concatenate the aligned multi-scale task features along the channel dimension:
\begin{equation}
X^{t}_{\mathrm{cat}} = \mathrm{Concat}\left(\tilde{X}^{t}_{1}, \tilde{X}^{t}_{2}, \tilde{X}^{t}_{3}, \tilde{X}^{t}_{4}\right)
\in \mathbb{R}^{4C \times H \times W}.
\end{equation}
We then employ a lightweight fusion head composed of a $1\times1$ convolution for channel mixing and compression, followed by batch normalization and a non-linear activation function, and a subsequent $3\times3$ convolution to capture local spatial interactions, corresponding to pixel-wise fusion logits for each stage:
\begin{equation}
L^{t} = \psi\!\left(X^{t}_{\mathrm{cat}}\right),
\qquad
L^{t}\in\mathbb{R}^{4\times H\times W},
\end{equation}
where $\psi(\cdot)$ denotes the convolutional fusion module.
% To aggregate task features across scales, we compute a task-dependent attention over stages.
% Specifically, we form a global descriptor $g^{t}_{i}=\mathrm{GAP}(\tilde{X}^{t}_{i})\in\mathbb{R}^{C}$ and predict a scalar logit:
% \begin{equation}
% \ell^{t}_{i}= \mathbf{w}^{\top}\sigma\!\left(\mathrm{MLP}(g^{t}_{i})\right).
% \end{equation}
% The normalized fusion weights are obtained by a softmax over stages:
% \begin{equation}
% a^{t}_{i} = \frac{\exp(\ell^{t}_{i})}{\sum_{j=1}^{4}\exp(\ell^{t}_{j})}, \qquad \sum_{i=1}^{4} a^{t}_{i}=1,
% \end{equation}
% and the multi-scale task feature is computed as the corresponding weighted sum:
% The obtained stage scores $\{\ell^{t}_{i}\}_{i=1}^{4}$ are then normalized to fusion weights and used to aggregate multi-scale features:
% \begin{equation}
% X^{t}_{\mathrm{agg}}=\sum_{i=1}^{4}\frac{\exp(\ell^{t}_{i})}{\sum_{j=1}^{4}\exp(\ell^{t}_{j})}\,\tilde{X}^{t}_{i}.
% \end{equation}
% \begin{equation}
% X^{t}_{\mathrm{agg}} = \sum_{i=1}^{4} a^{t}_{i}\,\tilde{X}^{t}_{i}.
% \end{equation}
A softmax operation is applied along the stage dimension to obtain normalized pixel-wise fusion weights:
\begin{equation}
A^{t}_{i}(h,w) = 
\frac{\exp\!\left(L^{t}_{i}(h,w)\right)}
{\sum_{j=1}^{4}\exp\!\left(L^{t}_{j}(h,w)\right)},
\qquad \sum_{i=1}^{4} A^{t}_{i}(h,w)=1.
\end{equation}
Finally, the aggregated task feature is computed as a pixel-wise weighted sum over scales:
\begin{equation}
X^{t}_{\mathrm{agg}}(h,w) = 
\sum_{i=1}^{4} A^{t}_{i}(h,w)\,\tilde{X}^{t}_{i}(h,w).
\end{equation}

In parallel, we merge semantic priors across stages into a single map.
Concretely, we concatenate $\{\tilde{S}_{i}\}_{i=1}^{4}$ along the channel dimension and apply a lightweight fusion operator:
\begin{equation}
S_{\mathrm{agg}} = \psi_{s}\!\left(\mathrm{Concat}\bigl(\tilde{S}_{1},\tilde{S}_{2},\tilde{S}_{3},\tilde{S}_{4}\bigr)\right),
\end{equation}
where $\psi_{s}(\cdot)$ is implemented as $1\times1$ channel mixing followed by a $3\times3$ convolution for local context integration.

% Given the aggregated task feature $X^{t}_{\mathrm{agg}}$, SPI injects semantic priors through feature-wise affine modulation.
% We first predict per-location scaling and bias terms from $S_{\mathrm{agg}}$ and then modulate the task feature as:
% \begin{equation}
% (\Gamma^{t},\Delta^{t})=\rho^{t}(S_{\mathrm{agg}}),
% \end{equation}
% \begin{equation}
% X^{t}_{\mathrm{mod}}=(1+\Gamma^{t})\odot X^{t}_{\mathrm{agg}}+\Delta^{t},
% \end{equation}
% where $\rho^{t}(\cdot)$ is a small convolutional predictor.
% The final prior-enhanced representation is produced by refinement:
% \begin{equation}
% Z^{t}=\psi_{z}\!\left(\mathrm{Concat}\bigl(X^{t}_{\mathrm{mod}},\,S_{\mathrm{agg}},\,X^{t}_{\mathrm{agg}}\bigr)\right),
% \end{equation}
% with $\psi_{z}(\cdot)$ realized by a Conv--Norm--GeLU--Conv stack.
Given the aggregated task feature $X^{t}_{\mathrm{agg}}$, SPI injects semantic priors by directly concatenating the aligned semantic representation with the task feature, followed by joint feature refinement.
Specifically, we fuse the task feature and semantic prior as:
\begin{equation}
U^{t} = \mathrm{Concat}\bigl(X^{t}_{\mathrm{agg}},\, S_{\mathrm{agg}}\bigr),
\end{equation}
where the concatenation is performed along the channel dimension.
The fused representation is then processed by a lightweight refinement block to produce the final prior-enhanced feature:
\begin{equation}
Z^{t} = \psi_{z}\!\left(U^{t}\right),
\end{equation}
where $\psi_{z}(\cdot)$ is implemented as a Conv--Norm--GeLU--Conv stack.

\subsubsection{Task-Specific Decoding Head}

Given $Z^{t}$, we employ a lightweight head that maps the shared latent to the prediction space of task $t$:
\begin{equation}
\hat{Y}^{t} = \mathcal{U}\bigl(\mathcal{P}^{t}(Z^{t})\bigr),
% \hat{Y}^{t} = \mathcal{U}\Bigl(\mathcal{P}^{t}\bigl(\mathrm{ConvBlock}(Z^{t})\bigr)\Bigr),
\end{equation}
where $\mathcal{P}^t(\cdot)$ is a task-dependent $1\times1$ projection and $\mathcal{U}(\cdot)$ denotes bilinear upsampling to the input resolution.

\subsubsection{Joint Optimization Losses}
\label{sec:3.3.3}
We train all tasks jointly with $\mathcal{L}_{\mathrm{total}}$, which is defined as:
\begin{equation}
\mathcal{L}_{\mathrm{total}}=\sum_{t=1}^{T}\lambda_{t}\,\mathcal{L}_{t},
\end{equation}
where $\mathcal{L}_{t}$ is task-specific (e.g., cross-entropy for segmentation, $\ell_{1}$/scale-invariant losses for depth, and cosine/angular losses for normals) and $\lambda_{t}$ balances tasks.
We use uniform weighting $\lambda_{t}=1$ in all experiments.

\begin{table*}
\centering
\caption{Performance comparison on the \textit{PASCAL-Context} datasets. $\uparrow$/$\downarrow$ indicates that the higher/lower results represent better performance. }
\label{Pascal Context Comparison}
\scalebox{1}{
\begin{tabular}{rlccccccccc}
% \hline
\toprule
% \cmidrule(lr){1-6} \cmidrule(lr){7-13} 
\multicolumn{2}{c}{\multirow{2}{*}{\textbf{Method}}} & \multirow{2}{*}{\textbf{Backbone}}& 
% \textbf{Params} & 
\textbf{Sem. Seg.} & \textbf{H. Pars.} & \textbf{Sal.} & \textbf{Norm.} & \textbf{Edge} & $\Delta_{m}$ & \textbf{Para.} & \textbf{Train. Para.} \\
& &  
% & (M) 
& mIoU$\uparrow$ & mIoU$\uparrow$ & maxF$\uparrow$ & mErr$\downarrow$ & odsF$\uparrow$ & $(\%)\uparrow$ & (M)$\downarrow$ & (M)$\downarrow$ \\
\cmidrule(lr){1-2}\cmidrule(lr){3-3}\cmidrule(lr){4-8}\cmidrule(lr){9-9}\cmidrule{10-11}
% \hline
% \cmidrule(lr){1-6}
 % Cross-Stitch& HRNet &&&&& ASTMT & HRNet &&&&&\\
 % PAD-Net~\cite{xu2018pad}&HRNet18& 50.23 & 0.5818 & - & - & PAD-Net~\cite{xu2018pad} & HRNet18 &&&&&\\
 % PSD&&&&&& MTI-Net &&&&&&\\
 % PAD-Net&&&&&& ATRC &&&&&&\\
 % STL & \!\!\pub{Ours}\!\! & Swin-L & 355 & 81.51 & 74.76 & 85.85 & 13.45 & 79.50 & 0.00 \\
MTL Baseline & \!\! \!\! & ViT-L 
% & 235 
& 82.61 & 76.38 & 77.09 & 17.63 & 74.00 & 0.00 & 328 & 25 \\
\cmidrule(lr){1-2}\cmidrule(lr){3-3}\cmidrule(lr){4-8}\cmidrule(lr){9-9}\cmidrule{10-11}
PAD-Net~\cite{xu2018pad} & \!\!\pub{CVPR2018}\!\! & HRNet-18 & 
% 81 & 
53.60 & 59.60 & 65.80 & 15.30 & 72.50& -12.11 & 81 & 81 \\
MTI-Net~\cite{vandenhende2020mti} & \!\!\pub{ECCV2020}\!\! & HRNet-18 & 
% 128 &
61.70 & 60.18 & 84.78 & 14.23 & 70.80& -4.32 & 128 & 128 \\
ATRC~\cite{bruggemann2021exploring} & \!\!\pub{ICCV2021}\!\! & HRNet-18 & 
% 96 &
67.67 & 62.93 &82.29 &14.24 &72.42& -2.37 & \underline{96} & 96 \\
ASTMT~\cite{maninis2019attentive}& \!\!\pub{ECCV2024}\!\!& ResNet-50 & 
% 49 & 
68.00 & 61.12 & 65.71 & 14.68 & 72.40 & -7.57 & \textbf{49} & \textbf{49} \\
\cmidrule(lr){1-2}\cmidrule(lr){3-3}\cmidrule(lr){4-8}\cmidrule(lr){9-9}\cmidrule{10-11}
% Cross-Stitch~\cite{misra2016cross} & \!\!\pub{ECCV2024}\!\! & ResNet-18 & & 66.12 & 60.66 & 66.81 & 13.89 & 69.90 & \\
% ASTMT~\cite{maninis2019attentive}& \!\!\pub{ECCV2024}\!\!& ResNet-50 & & 68.00 & 61.12 & 65.71 & 14.68 & 72.40 & \\
% MQTransformer~\cite{xu2023multi} & \!\!\pub{arXiv2022}\!\! & Swin-S & 55 & 71.25& 60.11& 84.05& 14.74& 71.80 & -6.20 \\
MTMamba~\cite{lin2024mtmamba}& \!\!\pub{ECCV2024}\!\!& Swin-L & 
% 308 & 
81.11 & 72.62 & 84.14 & 14.14 & \second{78.80} & 5.74 & 336 & 336 \\
MTMamba++~\cite{lin2025mtmamba++}& \!\!\pub{TPAMI2025}\!\!& Swin-L & 81.94 & 72.87 & \best{85.56} & 14.29 & 78.60 & 6.15 & 343 & 343 \\
BIM~\cite{cao2025enhancing} & \!\!\pub{ICCV2025}\!\!& Swin-L  & 81.25 & 73.15 & 84.14 & 14.13 & \best{79.05} & 5.99 & 388 & 388 \\
\cmidrule(lr){1-2}\cmidrule(lr){3-3}\cmidrule(lr){4-8}\cmidrule(lr){9-9}\cmidrule{10-11}
InvPT~\cite{ye2022inverted} & \!\!\pub{ECCV2022}\!\! & ViT-L & 
% 423 & 
79.03 & 67.61 & 84.81 & 14.15 & 73.00 & 2.52 & 423 & 423 \\
InvPT++~\cite{ye2024invpt++} &\!\!\pub{TPAMI2024}\!\! & ViT-L & 
% 421 & 
80.44 & 69.19 & 84.85 & 13.70 & 74.20 & 4.12 & 421 & 421 \\
TaskPrompter~\cite{ye2023taskprompter}& \!\!\pub{ICLR2023}\!\!& ViT-L & 
% 401 & 
80.89 & 68.89 & 84.83 & 13.72 & 73.50 & 3.93 & 401 & 401 \\
TaskExpert~\cite{ye2023taskexpert} & \!\!\pub{ICCV2023}\!\! & ViT-L &
% 420 & 
80.64 & 69.42 & 84.87 & \second{13.56} & 74.20 & 4.39 &  420 &  420 \\
MLoRE~\cite{yang2024multi} & \!\!\pub{CVPR2024}\!\! & ViT-L & 
% 407 & 
81.41 & 70.52 & 84.90 & \best{13.51} & 75.42 & 5.26  & 407 & 407\\
TaskDiffusion~\cite{yangmulti} & \!\!\pub{ICLR2025}\!\! & ViT-L & 81.21
% 416 & 
 & 69.62 & 84.94 & 13.55 & 74.89 & 4.80 &  416 &  416 \\
BridgeNet~\cite{zhang2025bridgenet} & \!\!\pub{TPAMI2025}\!\! & ViT-L & 80.64
% 477 & 
 & 70.06 & 84.64 & 13.82 & 72.96 & 3.87 & 477 & 477 \\
SAK~\cite{luswiss} & \!\!\pub{ICLR2025}\!\! & ViT-L & \second{84.01}
% 477 & 
 & \second{76.99} & 84.65 & 13.82 & 76.27 & \underline{7.40} & 407 & 407 \\
% \cdashline{1-11}
\ours& \!\!\pub{Ours}\!\! & ViT-L  & \best{84.48} & \best{78.05} & \second{85.55} & 13.57 & 78.40 & \textbf{8.88} & 358 & \second{54} \\
% \hline
 \bottomrule
\end{tabular}
}
\end{table*}
\begin{table*}
\centering
\caption{Performance comparison on the NYUDv2 dataset.
$\uparrow$/$\downarrow$ indicates that the higher/lower results represent better performance. }
\label{nyud comparison}
% \resizebox{\textwidth}{!}{
\scalebox{1.05}{
\begin{tabular}{rlcccccc}
% \hline
\toprule
% \multicolumn{6}{c}{\textbf{NYUD-v2}} \\
% \cmidrule(lr){1-6}
\multicolumn{2}{c}{\multirow{2}{*}{\textbf{Methods}}} & \multirow{2}{*}{\textbf{Backbone}} 
% & \textbf{Params} 
& \textbf{Sem. Seg.} &  \textbf{Dep.} & \textbf{Norm.} & \textbf{Edge} & $\Delta_{m}$ \\
& & 
% & (M) 
& mIoU$\uparrow$ & RMSE$\downarrow$ & mErr$\downarrow$ & odsF$\uparrow$ & $(\%)\uparrow$ \\
% \hline
\cmidrule(lr){1-2}\cmidrule(lr){3-3}\cmidrule(lr){4-7}\cmidrule(lr){8-8}
% \cmidrule(lr){1-6}  
 % Cross-Stitch& HRNet &&&&& ASTMT & HRNet &&&&&\\
 % PAD-Net~\cite{xu2018pad}&HRNet18& 50.23 & 0.5818 & - & - & PAD-Net~\cite{xu2018pad} & HRNet18 &&&&&\\
 % PSD&&&&&& MTI-Net &&&&&&\\
 % PAD-Net&&&&&& ATRC &&&&&&\\
 % STL & \!\!\pub{Ours}\!\! & Swin-L & 355 & 57.67 & 0.4769 & 18.05 & 79.2 & 0.0 \\
 MTL Baseline & \!\! \!\! & ViT-L 
 % & 235 
 & 63.15 & 0.5119 & 18.81 & 75.10 & 0.00 \\
 \cmidrule(lr){1-2}\cmidrule(lr){3-3}\cmidrule(lr){4-7}\cmidrule(lr){8-8}
 MTI-Net~\cite{vandenhende2020mti} & \!\!\pub{ECCV2024}\!\! & HRNet-18 
 % & 128 
 & 45.97 & 0.5365 & 20.27 & 77.84 & -9.03 \\
 PAD-Net ~\cite{xu2018pad} & \!\!\pub{CVPR2018}\!\! &HRNet-18 
 % & 81 
 & 36.61 & 0.6270 & 20.85 & 76.38 & -18.41 \\
 ATRC~\cite{bruggemann2021exploring} & \!\!\pub{ICCV2021}\!\! & HRNet-18 
 % & 96 
 & 46.33 & 0.5363 & 20.18 & 77.94 & -8.73 \\
 % Cross-Stitch~\cite{misra2016cross} & \!\!\pub{ECCV2024}\!\! & ResNet-50 & & 36.34 & 0.6290 & 20.88 & 76.38 & \\
 ASTMT~\cite{maninis2019attentive} & \!\!\pub{ECCV2024}\!\! & ResNet-50 
 % & 49 
 & 32.16 & 0.5700 & 23.18 & 74.50 & -21.11 \\
 \cmidrule(lr){1-2}\cmidrule(lr){3-3}\cmidrule(lr){4-7}\cmidrule(lr){8-8}
 MTMamba~\cite{lin2024mtmamba} & \!\!\pub{ECCV2024}\!\! & Swin-L 
 % & 308 
 & 55.82 & 0.5066 & 18.63 & 78.70& -1.21 \\
 MTMamba++~\cite{lin2025mtmamba++} & \!\!\pub{TPAMI2025}\!\! & Swin-L 
 % & 308 
 & 57.01 & 0.4818 & 18.27 & \underline{79.40} & 1.19 \\
 BIM~\cite{cao2025enhancing} & \!\!\pub{ICCV2025}\!\!& Swin-L  & 57.40 & 0.4733 & 18.55 & 78.72 & 1.16 \\
 \cmidrule(lr){1-2}\cmidrule(lr){3-3}\cmidrule(lr){4-7}\cmidrule(lr){8-8}
 InvPT~\cite{ye2022inverted} & \!\!\pub{ECCV2024}\!\! & ViT-L 
 % &  423 
 & 53.56 & 0.5183 & 19.04 & 78.10 & -3.42 \\
 InvPT++~\cite{ye2024invpt++} & \!\!\pub{ECCV2024}\!\!& ViT-L 
 % & 421 
 & 53.89 & 0.5091 & 18.65 & 78.10 & -2.32 \\
 TaskPrompter~\cite{ye2023taskprompter} & \!\!\pub{ECCV2024}\!\!& ViT-L 
 % & 401 
 & 55.30 & 0.5152 & 18.47 & 78.20 & -1.78 \\
 TaskExperts~\cite{ye2023taskexpert} & \!\!\pub{ECCV2024}\!\! & ViT-L 
 % & 420 
 & 55.35 & 0.5157 & 18.54 & 78.40 & -1.82 \\
 MLoRE~\cite{yang2024multi} & \!\!\pub{ECCV2024}\!\! & ViT-L 
 % & 407 
 & 55.96 & 0.5076 & 18.33 & 78.43 & -0.89 \\
 TaskDiffusion~\cite{yangmulti} & \!\!\pub{ICLR2025}\!\! & ViT-L 
 % & 407 
 & 55.65 & 0.5020 & 18.43 & 78.64 & -0.80 \\
 BridgeNet~\cite{zhang2025bridgenet} & \!\!\pub{TPAMI2025}\!\! & ViT-L 
 % & 407 
 & 55.51 & 0.4930 & 18.47 & 78.22 & -0.61 \\
 SAK~\cite{luswiss} & \!\!\pub{ICLR2025}\!\! & ViT-L 
 % & 407 
 & \underline{63.18} & \underline{0.4313} & \underline{16.25} & \underline{79.43} & \underline{8.79} \\
  % \cdashline{1-8}
 % MQTransformer~\cite{xu2023multi} & \!\!\pub{ECCV2024}\!\!& Swin-L & 204 & 54.84 & 0.5325 & 19.67 & 78.20 & -0.59 \\
\ours& \!\!\pub{Ours}\!\! & ViT-L  &  \textbf{65.91}&  \textbf{0.4209}&  \textbf{15.82}& \textbf{80.60} & \textbf{11.34} \\
 % \hline
 \bottomrule
\end{tabular}
}
\end{table*}

\section{Experiments}
\label{sec:4}
In this section, we evaluate \ours for multi-task dense prediction. \S\ref{sec4.1} introduces the benchmarks (Pascal-Context, NYUD-v2, and Cityscapes) and the evaluation protocol, including per-task metrics and the overall multi-task gain $\Delta_m$. \S\ref{sec4.2} details our training and implementation. \S\ref{sec4.3} reports quantitative and qualitative comparisons with recent state-of-the-art methods. \S\ref{sec4.4} presents ablations studies and analyze the contribution of each component. \S\ref{sec4.5} gives feature visualization and analysis. \S\ref{sec4.6} evaluates robustness.

\subsection{Experimental Setups.}
\label{sec4.1}
\subsubsection{Datasets}
Following recent MTL works~\cite{ye2024invpt++,zhang2025bridgenet,lin2024mtmamba}, we evaluate \ours  on three widely used benchmarks under standardized protocols, with official splits and their canonical task suites. \textbf{Pascal-Context}~\cite{chen2014detect} contains 4998 training and 5105 test images with rich pixel-level annotations, on which we report semantic segmentation (Sem. Seg.), human parsing (H. Pars.), object edge detection (Edge), surface normal estimation (Norm.), and saliency detection (Sal.). \textbf{NYUD-v2}~\cite{silberman2012indoor} focuses on indoor scenes with 795 training and 654 test samples; we evaluate Sem. Seg., monocular depth estimation (Dep.), Norm., and Edge \textbf{Cityscapes} provides urban street-view imagery with 2975 training images and 500 validation images, including 19 semantic classes and stereo-derived disparity; we follow common practice and report Sem. Seg. together with geometry-related predictions (Dep.) derived from the fine annotations. Together, these datasets cover outdoor scenes, indoor environments, and driving scenarios, enabling a broad and consistent evaluation of multi-task dense prediction under diverse conditions.

% Following recent MTL work~\cite{ye2024invpt++,zhang2025bridgenet,lin2024mtmamba}, we evaluate \ours  on two standard benchmarks using official splits and canonical task suites. Pascal-Context~\cite{chen2014detect} provides 4998 train and 5{,}105 test images with pixel-level annotations; we report semantic segmentation (SemSeg), human parsing (Parsing), object boundary detection (Boundary), surface normal estimation (Normal), and saliency detection (Saliency). NYUD-v2~\cite{silberman2012indoor} comprises indoor scenes with 795 train and 654 test samples; we evaluate semantic segmentation, monocular depth estimation (Depth), surface normal estimation (Normal), and object boundary detection (Boundary). Together, these datasets span outdoor/general and indoor environments, enabling consistent, multi-task dense prediction comparisons.

\subsubsection{Evaluation Metrics}
To enable a rigorous, fair and fully comprehensive comparison for multi-task dense prediction, we adopt standard task-specific measures and a normalized cross-task summary metric for overall assessment. Specifically, we report:
\begin{itemize}[leftmargin=1.6em,labelsep=0.6em]
\item \textbf{Sem. Seg.} / \textbf{H. Pars.}: mean Intersection-over-Union (mIoU).
\item \textbf{Dep.}: root mean square error (RMSE).
\item \textbf{Norm.}: mean angular error (mErr).
\item \textbf{Sal.}: maximal F-measure (maxF).
\item \textbf{Edge}: Optimal Dataset Scale F-measure (odsF).
\end{itemize}

\begin{table}
\centering
\caption{Performance comparison on the Cityscapes dataset.}
% with state-of-the-art methods
% $\uparrow$/$\downarrow$ indicates that the higher/lower results represent better performance. }
\label{cityscape comparison}

\begin{tabular}{lcccc}
\toprule
\multirow{2}{*}{\textbf{Methods}} & \multirow{2}{*}{\textbf{Backbone}} 
& \textbf{Sem. Seg.} &  \textbf{Dep.} & $\Delta_{m}$ \\
& & mIoU$\uparrow$ & RMSE$\downarrow$ & $(\textit{\%})\uparrow$ \\
\cmidrule(lr){1-1}\cmidrule(lr){2-2}\cmidrule(lr){3-4}\cmidrule(lr){5-5}

% ======================
%   OUR BASELINE
% ======================
MTL Baseline & ViT-L 
& 76.45 & 4.36 & 0 \\
\cmidrule(lr){1-1}\cmidrule(lr){2-2}\cmidrule(lr){3-4}\cmidrule(lr){5-5}

% ======================
%   HRNet-18 BLOCK
% ======================
MTI-Net~\cite{vandenhende2020mti} 
& HRNet-18 
& 53.19 & 5.05 & -23.13 \\
PAD-Net~\cite{xu2018pad} 
& HRNet-18 
& 59.85 & 5.06 & -18.88 \\
\cmidrule(lr){1-1}\cmidrule(lr){2-2}\cmidrule(lr){3-4}\cmidrule(lr){5-5}

% ======================
%   Swin-L BLOCK
% ======================
MTMamba~\cite{lin2024mtmamba} 
& Swin-L 
& 78.00 & 4.66 & -2.43 \\
MTMamba++~\cite{lin2025mtmamba++} 
& Swin-L 
& 79.13 & 4.63 & -1.34 \\
\cmidrule(lr){1-1}\cmidrule(lr){2-2}\cmidrule(lr){3-4}\cmidrule(lr){5-5}

% ======================
%   ViT-L BLOCK
% ======================
InvPT~\cite{ye2022inverted} 
& ViT-L 
& 71.78 & 4.67 & -6.61 \\
TaskPrompter~\cite{ye2023taskprompter} 
& ViT-L 
& 72.41 & 5.49 & -15.60 \\
% \cdashline{1-5}

% ======================
%   OUR METHOD
% ======================
\ours
& ViT-L  
& \textbf{82.10} & \textbf{4.22} & \textbf{5.30} \\
\bottomrule
\end{tabular}
\end{table}

To summarize performance across heterogeneous task objectives, the overall performance gain of multi-task learning ($\Delta_{m}$) is computed relative to our multi-task baseline:
\begin{equation}
\Delta_{m}=\frac{1}{T}\sum_{t=1}^{T}(-1)^{\ell_{t}}\frac{M_{t}-M^{m}_{t}}{M^{m}_{t}},
\end{equation}
where $T$ is the number of tasks, $M_{t}$ and $M^{m}_{t}$ denote the comparing result and our multi-task baseline results for task $t$, and $\ell_{t}=1$ for lower-is-better metrics (RMSE, mErr) and $\ell_{t}=0$ otherwise (mIoU, maxF, ODS-F). A positive $\Delta_{m}$ therefore indicates an overall improvement. We report all per-task metrics together with $\Delta_{m}$ on the official dataset splits.
% To ensure a fair and comprehensive comparison in multi-task dense prediction, we follow established practice by using task-specific metrics together with a normalized cross-task summary. For semantic segmentation and human parsing we report mean Intersection-over-Union (mIoU); for monocular depth/disparity estimation, root mean square error (RMSE); for surface normal estimation, mean angular error (mErr); for saliency detection, maximal F-measure (maxF); and for object boundary detection, the Optimal Dataset Scale F-measure (ODS-F). To aggregate effectiveness across heterogeneous goals, we compute the multi-task learning gain ($\Delta_{m}$) relative to single-task baselines:
% \begin{equation}
% \Delta_{m}=\frac{1}{T}\sum_{t=1}^{T}(-1)^{\ell_{t}}\frac{M^{m}_{t}-M^{s}_{t}}{M^{s}_{t}},
% \end{equation}
% where $T$ is the number of tasks, $M^{m}_{t}$ and $M^{s}_{t}$ are the multi-task and single-task scores on task $t$, and $\ell_{t}=1$ for lower-is-better metrics (RMSE, mErr) and $0$ otherwise (mIoU, maxF, ODS-F). Positive $\Delta_{m}$ indicates an overall improvement, and we report both per-task metrics and $\Delta_{m}$ on the official splits.
\begin{figure*}[t]
    \centering % 0.8 scale(double c)
    \includegraphics[width=0.8\linewidth]{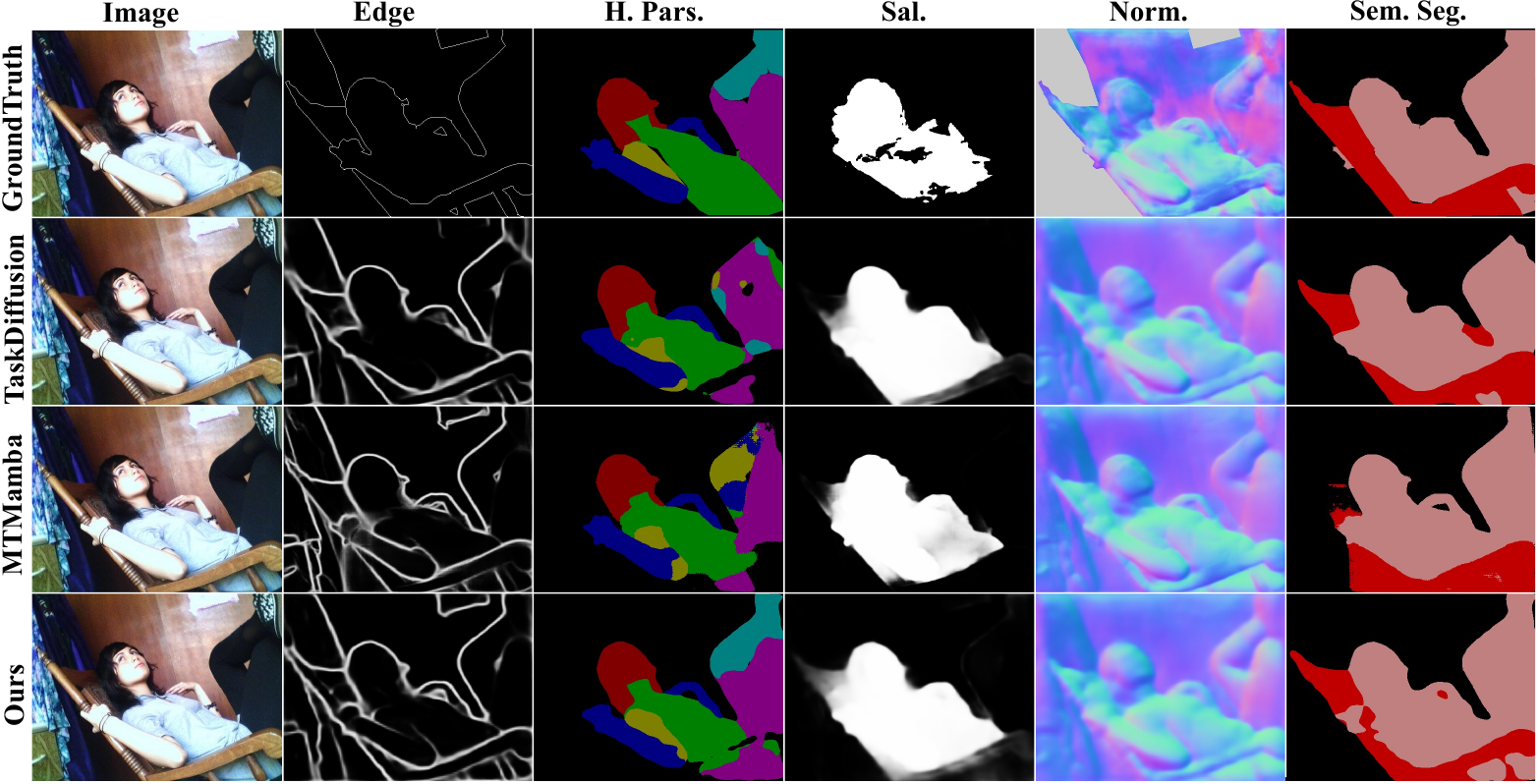}
    \caption{\textbf{Qualitative Comparison on Pascal-Context.}
Compared with TaskDiffusion~\cite{yangmulti} and MTMamba~\cite{lin2024mtmamba}, \ours produces crisper boundaries and fewer cross-task inconsistencies (e.g., aligned silhouettes and object extents across parsing/segmentation/saliency and normals), indicating improved shared structure modeling and reduced task interference in multi-objective dense prediction.}
% our method yields sharper object boundaries and more stable, task-consistent predictions across semantic segmentation, human parsing, saliency detection, surface normal estimation, and edge detection.}
    \label{fig:pascal_vis} 
\end{figure*} 

\subsection{Implementation Details}
\label{sec4.2}
{
\looseness=-1
To enable parameter-efficient multi-task dense prediction, we use DINOv3~\cite{simeoni2025dinov3} as frozen backbone and adapt it with our designed modules. 
Models are trained with AdamW (weight decay $1\times10^{-6}$) and a polynomial learning rate schedule. Base learning rates are $8\times10^{-5}$, $2\times10^{-5}$, and $1\times10^{-4}$ for Pascal-Context, NYUDv2, and Cityscapes, respectively. We train on two NVIDIA A100 GPUs with a batch size of $4$ ($2$ per GPU). Loss functions are detailed in Sec.~\ref{sec:3.3.3}. TG-LoRA is applied to the $Q$, $K$, $V$, and output layers of all attention blocks with rank $16$. TA-MoE is applied in the final attention layer, with each expert as a rank-$4$ LoRA module. The SPI module has a fused feature embedding dimension of 384. All LoRA adapters use a scaling factor of $4$ and a dropout rate of $0.05$. 
\par
}

To isolate the effect of our modules from DINOv3 itself, we build the MTL Baseline with the same frozen DINOv3 backbone and identical training protocol as \ours. This baseline trains only task-specific prediction heads, without TG-LoRA, TA-MoE, or SPI. Thus, its gap to \ours directly reflects the contribution of our proposed modules.
% \textcolor{blue}{To isolate the effect of the proposed adaptation modules from the DINOv3 backbone, we construct an MTL Baseline using the same frozen DINOv3 backbone as \ours. This baseline keeps the entire DINOv3 backbone frozen and only equips it with task-specific prediction heads for different dense prediction tasks, without using TG-LoRA, TA-MoE, or SPI. Therefore, the comparison between the MTL Baseline and \ours directly reflects the contribution of the proposed adaptation modules under the same backbone setting. All models }% R1_C2

\subsection{Benchmark Results}
\label{sec4.3}

\subsubsection{Quantitative Comparison Results}
We conduct comprehensive quantitative comparisons on three standard benchmarks, namely \textit{Pascal-Context}, \textit{NYUD-v2}, and \textit{Cityscapes}, using the official data splits and widely adopted multi-task settings for fair evaluation. We report task-specific metrics for each benchmark and additionally summarize overall performance using the aggregated multi-task gain $\Delta_m$, which facilitates a holistic comparison across heterogeneous dense prediction tasks.

\textbf{Pascal-Context.}
% \subsubsection{Quantitative Results on Pascal-Context}
Tab.~\ref{Pascal Context Comparison} summarizes results on \textit{PASCAL}\textit{-Context} across five dense prediction tasks including semantic segmentation (Sem. Seg.), human parsing (H. Pars.), saliency detection (Sal.), surface normal estimation (Norm.) and edge detection (Edge) under the standard multi-task setting. \ours\ achieves the highest aggregated multi-task gain ($\Delta_m=8.88$). Relative to the \textit{MTL Baseline} (ViT-L), it improves Sem. Seg. ($82.61$$\rightarrow$$84.48$ mIoU), H. Pars. ($76.38$$\rightarrow$$78.05$ mIoU) and Sal. ( $77.09$$\rightarrow$$85.55$ maxF), while reducing Norm. error ($17.63$$\rightarrow$$13.57$ mErr) and increasing boundary accuracy ($74.00$$\rightarrow$$78.40$ odsF). The concurrent gains across recognition-oriented and geometry-sensitive tasks suggest that \ours\ mitigates negative transfer while preserving beneficial sharing. Compared with decoder-centric methods such as MTMamba++ and BIM (Swin-L), \ours\ attains higher performance on segmentation and parsing and a larger overall gain ($8.88$ versus $6.15$ and $5.99$). Among ViT-L-based approaches, \ours\ also improves upon MLoRE, TaskDiffusion and BridgeNet, consistent with the view that task-adaptive routing, coupled with the semantic priors provided by DINOv3, supports stronger task disentanglement and more effective cross-task coordination.

\textbf{NYUD-v2.}
% \subsubsection{Quantitative Results on NYUD-v2}
Tab.~\ref{nyud comparison} reports quantitative results on NYUD-v2 for Sem. Seg, Dep., Norm. and Edge \ours\ achieves the highest aggregated multi-task gain ($\Delta_m=11.34$), with improvements observed across all tasks. Relative to the MTL baseline (ViT-L), performance increases from $63.15$ to $65.91$ mIoU for Sem. Seg and from $75.10$ to $80.60$ odsF for Edge, while Dep. RMSE is reduced from  $0.5119$ to $0.4209$ and Norm. from $18.81$ to $15.82$ mErr. Compared with recent strong baselines (\eg, BIM, TaskDiffusion), \ours\ shows a larger overall gain, with particularly pronounced benefits on geometry-related predictions, consistent with the importance of modelling indoor structure and fine-grained details.

\begin{figure*}[t]
    \centering % 0.8 scale(double c)
    \includegraphics[width=0.7\linewidth]{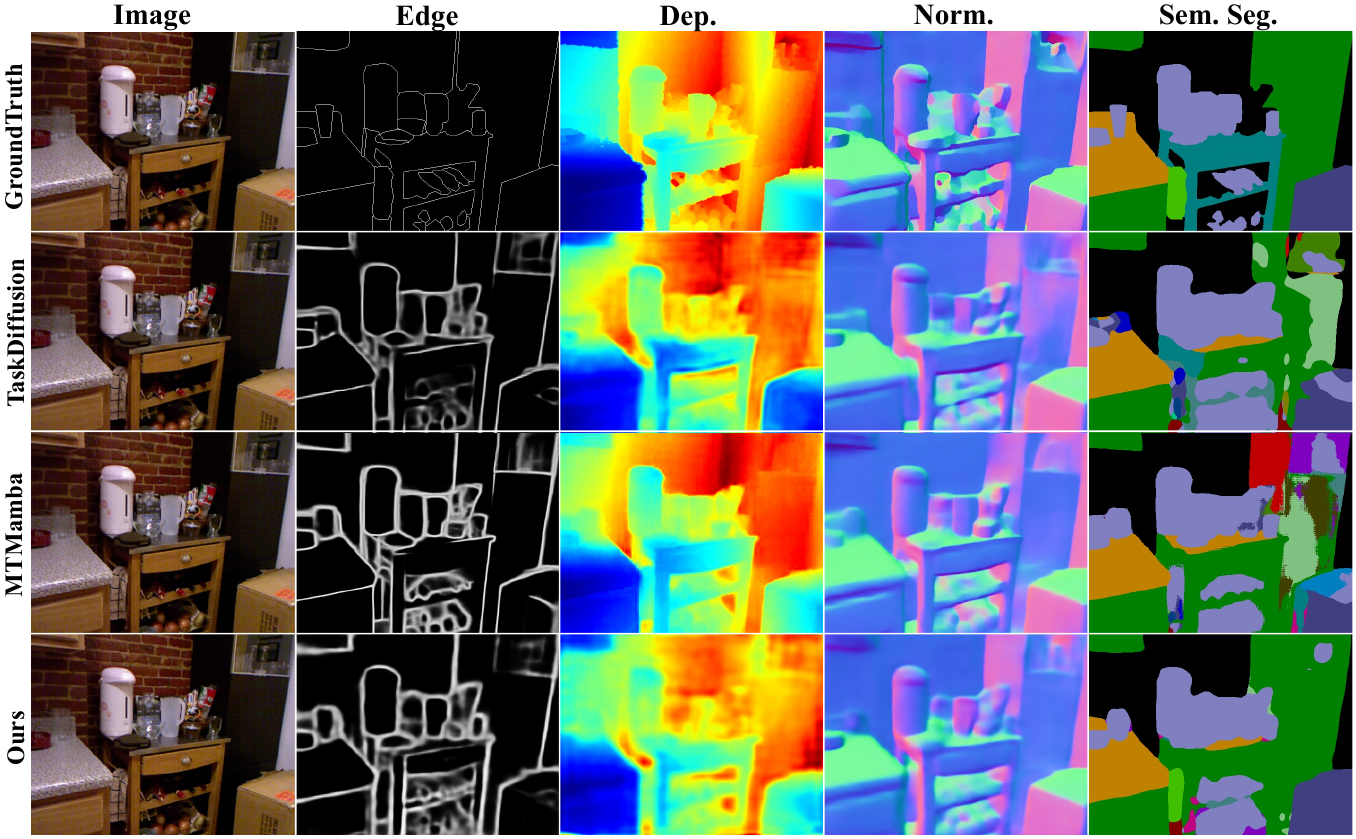}
    \caption{\textbf{Qualitative Comparison on NYUD-v2}. Compared with TaskDiffusion~\cite{yangmulti} and MTMamba~\cite{lin2024mtmamba}, \ours better preserves depth/normal discontinuities at object boundaries and yields structurally coherent geometry that agrees with segmentation, suggesting stronger cross-task geometric consistency across heterogeneous dense objectives.}
    % Visual results across edge detection, depth estimation, surface normal estimation and semantic segmentation compared against TaskDiffusion~\cite{yangmulti} and MTMamba~\cite{lin2024mtmamba}. 
    % \ours consistently yields superior boundary fidelity and structural coherence across heterogeneous dense objectives.}
    % \textbf{Visual comparison comparing to TaskDiffusion~\cite{yangmulti} and MTMMamba~\cite{lin2024mtmamba} on the NYUD-v2 dataset.}
% Our method consistently yields cleaner object boundaries and more stable structural cues across all tasks.}
    \label{fig:nyu_vis} 
\end{figure*} 
\begin{figure*}[t]
    \centering % 0.8 scale(double c)
    \includegraphics[width=1\linewidth]{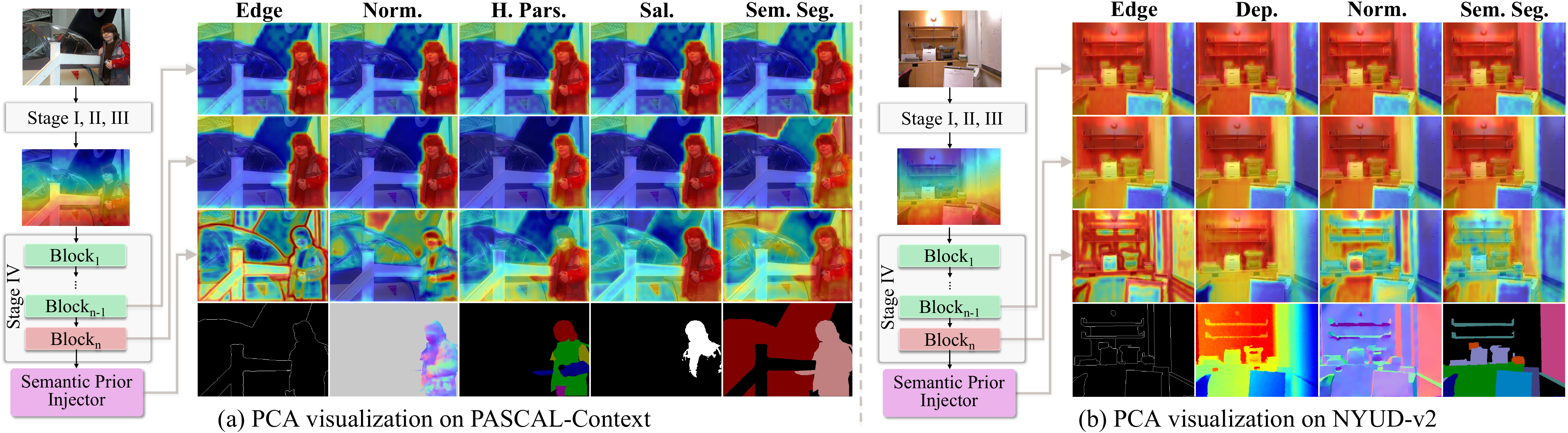}
    \caption{\textbf{PCA Visualizations on Pascal-Context (Left) and NYUD-v2 (Right).}
PCA visualizations are provided to analyze feature representations learned with TA-MoE, TG-LoRA, and the Semantic Prior Injector (SPI). As shown, features extracted from different network stages exhibit progressively clearer task-specific structures, demonstrating the effectiveness of the proposed modules in enhancing task-aware representation learning across both datasets.}
    \label{fig:pascal_pca_vis} 
\end{figure*} 

\textbf{Cityscapes.}
% \subsubsection{Quantitative Results on Cityscapes}
We further evaluate \ours\ on Cityscapes for joint Sem. Seg. and Dep. (Tab.~\ref{cityscape comparison}). \ours\ attains $82.10$ mIoU and $4.22$ RMSE, corresponding to a $\Delta_m=5.30$ overall gain with the ViT-L baseline ($76.45$ mIoU and $4.36$ RMSE). Notably, several existing approaches yield limited or negative overall gains under the same protocol, reflecting the optimisation trade-offs between semantic recognition and geometric estimation in urban driving scenes.
We further observe that the relative advantage of \ours is more evident on Cityscapes than on NYUD-v2 and Pascal-Context. This may stems from its regular driving layouts and tight semantic--geometric coupling, where roads, vehicles, and buildings follow stable spatial/depth patterns, making DINOv3 priors and task-aware routing more effective.
% \textcolor{blue}{We observe that \ours achieves more pronounced gains on Cityscapes than on NYUD-v2 and Pascal-Context. This is mainly because Cityscapes contains structured driving scenes with strong spatial regularities and cross-task consistency, allowing \ours to better exploit DINOv3 semantic priors and task-aware routing. In contrast, NYUD-v2 involves cluttered indoor scenes with stronger occlusion and scale variation, while Pascal-Context contains more diverse object-centric scenes with weaker shared spatial layouts. Therefore, the gains on these two datasets are relatively moderate but still consistent.}% R2_C3

Across Pascal-Context, NYUD-v2 and Cityscapes, \ours\ consistently yields the largest aggregated multi-task gain $\Delta_m$, despite substantial differences in visual domains and task compositions. This pattern strongly supports our motivation that effective MTDP with a frozen foundation backbone requires (i) \emph{task-generic} adaptation to align pretrained representations with dense prediction, and (ii) \emph{sparse, task-adaptive} specialization to resolve conflicts. 

\subsubsection{Qualitative Comparison}
\label{sec4.3.2}
We present qualitative results on Pascal-Context and NYUD-v2 to visually assess \ours. Fig.~\ref{fig:pascal_vis} and Fig.~\ref{fig:nyu_vis} compare \ours with strong recent baselines (TaskDiffusion and MTMamba) under the same multi-task setting.

\textbf{Pascal-Context.}
As shown in Fig.~\ref{fig:pascal_vis}, \ours yields more accurate and spatially coherent predictions across tasks. For Sem. Seg., \ours better preserves object extent and thin structures (\eg, limbs), reducing the label fragmentation observed in competing methods. For Norm., \ours produces smoother orientation fields with sharper transitions at geometric discontinuities, indicating stronger local shape consistency. For Edge, \ours outputs sharper and more continuous contours aligned with object silhouettes. For Sal., \ours improves foreground--background separation with fewer spurious activations. 

\textbf{NYUD-v2.} Fig.~\ref{fig:nyu_vis} demonstrates gains in indoor scene parsing. \ours yields Depth maps with refined geometry and reliable occlusion ordering, reducing planar noise. Norm. maps exhibit spatial consistency with crisp boundary transitions. For Sem. Seg., \ours successfully disambiguates adjacent objects via DINOv3 semantic priors. Finally, Edge results show improved completeness and connectivity even in cluttered environments.

\begin{figure*}[t]
    \centering % 0.8 scale(double c)
    \includegraphics[width=1\linewidth]{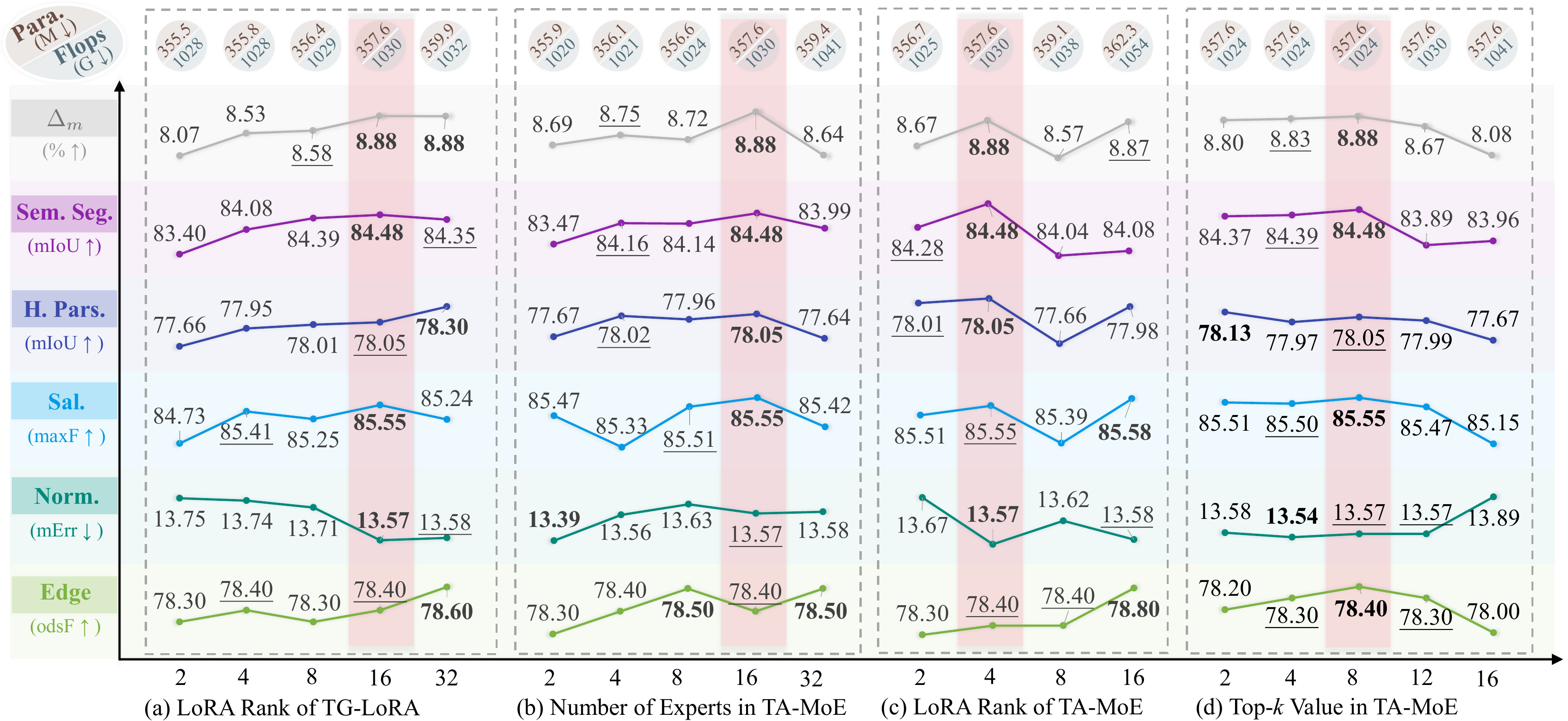}
    \caption{\textbf{Hyperparameter analysis of TG-LoRA and TA-MoE.}
The results demonstrate clear performance trends across different LoRA ranks and expert configurations, highlighting an effective trade-off between model capacity and multi-task performance.
}
    \label{fig:hyperparameters} 
\end{figure*}

\subsection{Ablation Study}
\label{sec4.4}
We conduct comprehensive ablation studies on the Pascal-Context dataset. In particular, \S\ref{sec4.4.1} isolates the contributions of our proposed modules and their placement, whereas \S\ref{sec4.4.2} examines the key hyperparameters that operationalize the intended balance between conditional specialization and parameter efficiency.

\subsubsection{Component and Design Ablations}
\label{sec4.4.1}
\textbf{Overall Components Analysis.}
Tab.~\ref{overall_ablation} disentangles the contributions of the Mixture of Task Adapters (MoTA) and Semantic Prior Injector (SPI) to our prior-preserving multi-task framework. Relative to the MTL baseline, MoTA alone increases the aggregated gain from $0.00$ to $6.10\%$, with marked improvements in Sal. ($82.61 \rightarrow 84.12$ mIoU) and Norm. ($15.60 \rightarrow 14.01$ mErr), supporting the need for task-conditioned adaptation to separate dense objectives without perturbing the frozen backbone. SPI alone provides a smaller gain ($\Delta_m = 5.66\%$), implying that semantic priors are insufficient to resolve cross-task interference under a single shared adaptation path. Enabling both yields the best performance, \ie $84.48$ mIoU (segmentation), $78.05$ mIoU (parsing), $85.55$ maxF (saliency), and $\Delta_m = 8.88\%$, consistent with complementarity: MoTA supplies controlled specialization, whereas SPI re-injects frozen semantic structure to improve global coherence and boundary fidelity.

\textbf{Ablation Results of MoTA.}
Tab.~\ref{moe_ablation} probes MoTA’s internal factorization. Removing TG-LoRA lowers the gain to $7.74\%$, indicating degradation of the task-generic alignment channel that stabilizes downstream adaptation; removing TA-MoE produces a comparable reduction ($\Delta_m = 7.97\%$), reflecting weaker task-specific discrimination without conditional expert routing. The full model outperforms both variants across tasks, showing that frozen-backbone multi-task learning benefits from jointly coupling a globally shared low-rank substrate (TG-LoRA) with sparse task-adaptive specialization (TA-MoE) rather than relying on either uniform adaptation or specialization alone.

\begin{table}
\centering
\caption{Overall ablation of the proposed components (MoTA and SPI) under the unified multi-task setting.}
\vspace{-2mm}
\label{overall_ablation}
\scalebox{0.85}{
% \resizebox{\columnwidth}{!}{
\begin{tabular}{cccccccccc}
\toprule
 \multirow{2}{*}{\textbf{MoTA}} & \multirow{2}{*}{\textbf{SPI}} & \textbf{Sem. Seg.} & \textbf{H. Pars.} & \textbf{Sal.} & \textbf{Norm.} & \textbf{Edge} & $\Delta_{m}$ \\
 & & mIoU$\uparrow$ & mIoU$\uparrow$ & maxF$\uparrow$ & mErr$\downarrow$ & odsF$\uparrow$ & $(\textit{\%})\uparrow$ \\
\cmidrule(lr){1-2}\cmidrule(lr){3-7}\cmidrule(lr){8-8}
% \multicolumn{2}{c}{MTL Baseline} & 82.61 & 76.38 & 77.09 & 17.63 &  & 0.00 \\
 % \hdashline
 % \cdashline{1-8}
 \ding{55} & \ding{55} & 82.61 & 76.38 & 77.09 & 17.63 & 74.00 & 0.00  \\
 \ding{55} & \ding{51} & \underline{83.27} & \underline{77.21} & 82.51 & 15.24 & \underline{78.30} & 5.66 \\
 \ding{51} & \ding{55} & 82.44 & 74.90 & \underline{84.12} & \underline{14.01} & 76.20 & \underline{6.10} \\
 \ding{51} & \ding{51} & \textbf{84.48} & \textbf{78.05} & \textbf{85.55} & \textbf{13.57} & \textbf{78.40} & \textbf{8.88} \\
\bottomrule
\end{tabular}
}
\end{table}

\begin{table}
\centering
\caption{Component ablation of Mixture of Task Adapters (MoTA), isolating the contributions of TG-LoRA and TA-MoE under the unified multi-task setting.}
% Ablation study of the Mixture of Task Adapters components, evaluating the contributions of TG-LoRA and TA-MoE under a unified multi-task setting.}
\vspace{-2mm}
\label{moe_ablation}
\scalebox{0.85}{
\begin{tabular}{cccccccc}
\toprule
 \multirow{2}{*}{\textbf{Modules}} & \textbf{Sem. Seg.} & \textbf{H. Pars.} & \textbf{Sal.} & \textbf{Norm.} & \textbf{Edge} & $\Delta_{m}$   \\
 & mIoU$\uparrow$ & mIoU$\uparrow$ & maxF$\uparrow$ & mErr$\downarrow$ & odsF$\uparrow$ & $(\textit{\%})\uparrow$ \\
\cmidrule(lr){1-1}\cmidrule(lr){2-6}\cmidrule(lr){7-7}
MTL Baseline & 82.61 & 76.38 & 77.09 & 17.63 & 74.00 & 0.00 \\
% \cdashline{1-7}
w/o TG-LoRA & 84.07 & \underline{77.92} & 84.32 & 14.15 & \textbf{78.30} & 7.74 \\
w/o TA-MoE & \underline{83.31} & 77.60 & \underline{84.35} & \textbf{13.65} & 78.00 & \underline{7.97} \\
Ours& \textbf{84.48} & \textbf{78.05} & \textbf{85.55} & \underline{13.57} & \underline{78.40} & \textbf{8.88} \\
 \bottomrule
\end{tabular}
}
\end{table}

\begin{table}[!t]
\centering
\large
\caption{Effect of inserting the MoE module at different locations within the backbone transformer block.}
% Impact of inserting the MoE modules at different locations within the transformer blocks of backbone.}
\vspace{-2mm}
\label{moe_location}
% \resizebox{\columnwidth}{!}{
\scalebox{0.58}{
\begin{tabular}{cccccccccc}
\toprule
 \multirow{2}{*}{\textbf{QKV}} & \multirow{2}{*}{\textbf{Proj}} & \multirow{2}{*}{\textbf{FFN}} & \textbf{Sem. Seg} & \textbf{H. Pars.} & \textbf{Sal.} & \textbf{Norm.} & \textbf{Edge} & $\Delta_{m}$ \\
 & & & mIoU$\uparrow$ & mIoU$\uparrow$ & maxF$\uparrow$ & mErr$\downarrow$ & odsF$\uparrow$ & $(\textit{\%})\uparrow$ \\
\cmidrule(lr){1-3}\cmidrule(lr){4-8}\cmidrule(lr){9-9}
% MTL Baseline & 82.61 & 76.38 & 77.09 & 17.63 & 74.00 & 0.00 \\
% \cdashline{1-7}
\ding{55} & \ding{55} & \ding{55} & 83.31 & 77.60 & 84.35 & 13.65 & 78.00 & 7.97 \\
\ding{51} & \ding{55} & \ding{55} & 83.74 & \underline{77.80} & \underline{85.18} & \underline{13.57} & 78.70 & 8.62 \\
\ding{55} & \ding{51} & \ding{55} & 83.31 & 77.67 & 84.86 & 13.73 & \underline{78.80} & 8.24 \\
\ding{55} & \ding{55} & \ding{51} & 83.34 & 77.24 & 84.63 & 13.58 & 78.00 & 8.03 \\
\ding{51} & \ding{51} & \ding{55} & \textbf{84.48} & \textbf{78.05} & \textbf{85.55} & \underline{13.57} & 78.40 & \textbf{8.88} \\
\ding{51} & \ding{51} & \ding{51} & \underline{83.75} & 77.51 & 85.02 & \textbf{13.51} & \textbf{79.10} & \underline{8.68} \\
 \bottomrule
\end{tabular}
}

\vspace{2pt}
  {\footnotesize\raggedright $\Delta_{m}$ s computed relative to the MTL baseline, which is omitted. \par}
\vspace{-4pt}
\end{table}

\textbf{Ablation Results of TA-MoE.}
Tab.~\ref{moe_location} evaluates where specialization most effectively alleviates the single-route bottleneck. Injecting TA-MoE into attention QKV projections outperforms insertion solely in the FFN or projection layers, indicating that conditioning token mixing (task-dependent contextual aggregation) is more effective than modifying downstream feed-forward transformations. Combining QKV and projection layers achieves the maximum gain ($8.88\%$), suggesting that co-conditioning attention and output projection best supports task disentanglement.

\textbf{Backbone Variants.}
Tab.~\ref{Backbone Ablation} quantifies task-dependent scaling across ViT backbones. ViT-S to ViT-B gives the largest segmentation jump ($80.62 \rightarrow 85.25$ mIoU) with modest parameter growth.
% , indicating that stronger shared features benefit dense recognition.
Scaling to ViT-L shifts gains to structured prediction: H. Pars. $75.66 \rightarrow 78.05$ mIoU, Sal. and Norm. peak at $85.55$ maxF and $13.57$ mErr, while Sem. Seg. saturates at $84.48$ mIoU. ViT-H adds slightly for H. Pars./Edge ($78.87$ mIoU, $79.60$ odsF) but degrades Sal. ($83.17$ maxF) and Norm. ($13.62$ mErr), implying diminishing or negative returns from additional backbone capacity. This degradation likely arises because the larger ViT-H encodes stronger high-level semantic invariances, while our frozen-backbone and low-rank adaptation cannot fully recalibrate its spatial-frequency bias for saliency and normals, making these geometry-sensitive tasks prone to over-smoothing. Overall, ViT-L provides the optimal accuracy-efficiency trade-off.

\textbf{Comparison with MLoRE under a Strong Backbone.}
Tab.~\ref{compare_mlore} reimplements MLoRE on DINOv3. MLoRE places experts in the decoder, improving task outputs over the MTL baseline and substantiating decoder-side specialization. D$^2$MoE instead instantiates experts in the backbone, decoupling tasks during representation learning while retaining DINOv3 priors through conditional adaptation. This shift delivers larger mean gains ($\Delta_m{=}8.88\%$ vs.\ $6.30\%$) with substantially fewer trainable parameters ($54$M vs.\ $407$M). Although \ours incurs higher FLOPs/latency than MLoRE (1029.9G/224.2ms vs.\ 829.9G/203.2ms), the additional computation is moderate and leads to stronger multi-task performance with a $7.5\times$ smaller trainable footprint, demonstrating a favorable trade-off among accuracy, trainable footprint, and computational cost.

\textbf{Comparison with Other Variants under Matched Budgets.}
Tab.~9 compares three parameter- and protocol-matched controls under the same frozen DINOv3 setting. Shared LoRA uses a single task-shared LoRA ($r=171$) at the same QKV/projection positions, achieving $\Delta_m=8.48\%$. Task LoRA replaces routing with independent task-wise LoRA modules ($r=186$), yielding $\Delta_m=8.31\%$. Decoder MoE moves the low-rank MoE to the decoder, using the same expert pool and Top-$8$ routing with 25.2M MoE parameters, but reaches only $\Delta_m=5.90\%$. In comparison, D$^2$MoE achieves the best $\Delta_m=8.88\%$. Overall, these matched controls show that the gains of D$^2$MoE are not explained by parameter count alone, but arise from combining shared low-rank alignment with sparse, learned task-aware specialization at appropriate backbone locations.

\begin{figure*}
    \centering
    \includegraphics[width=0.95\linewidth]{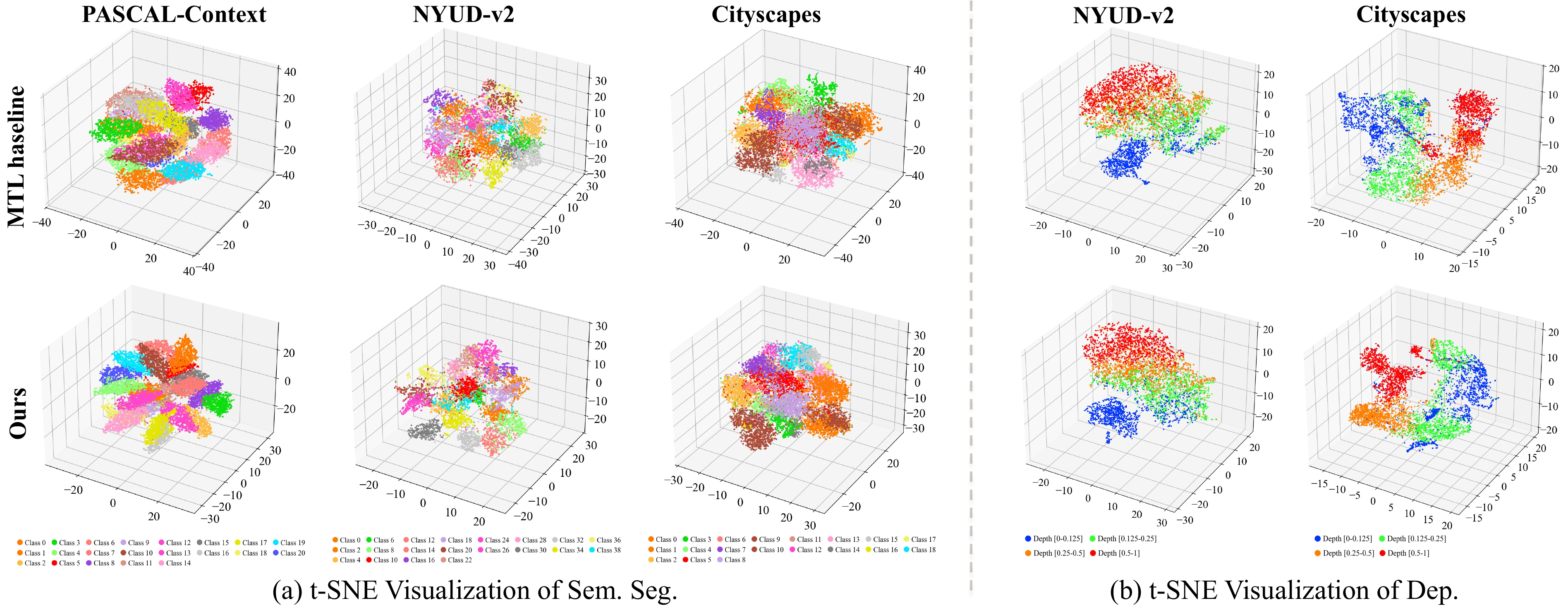}
    \caption{\textbf{3D t-SNE Visualizations of Feature Embeddings}. We compare the MTL baseline (top) with our \ours framework (bottom) across three benchmarks. (a) Semantic Segmentation: Points are colored by class label. \ours achieves significantly higher intra-class compactness and inter-class separability than the baseline. (b) Depth Estimation: Points are colored by discretized depth intervals (see legend). Our model yields more structured, monotonic groupings of scene geometry. Axes denote the three primary t-SNE components.}
    \label{fig:tsne_semseg}
\end{figure*}

\begin{figure*}
    \centering
    \includegraphics[width=0.98\linewidth]{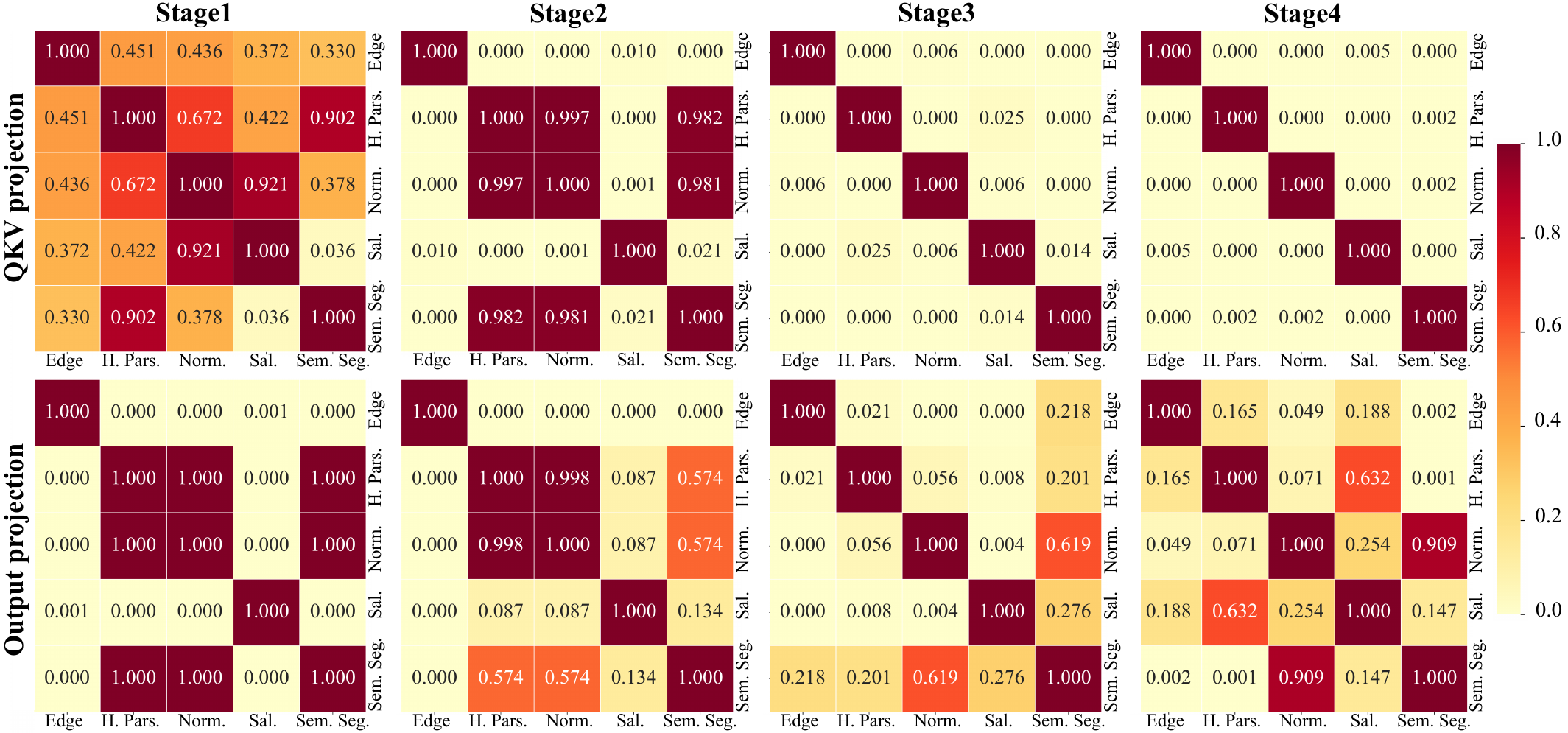}
    \caption{\textbf{Router visualization of TA-MoE across backbone stages.}
We visualize the task-wise expert selection patterns of TA-MoE inserted at different backbone stages and positions. For each task, we aggregate expert activation values over all samples in the Pascal-Context dataset and compute the cosine similarity between task-specific activation vectors. The resulting similarity matrices reveal how task relationships evolve across stages and insertion locations.}
    \label{fig:router_vis}
\end{figure*}

\begin{figure*}
    \centering
    \includegraphics[width=1\linewidth]{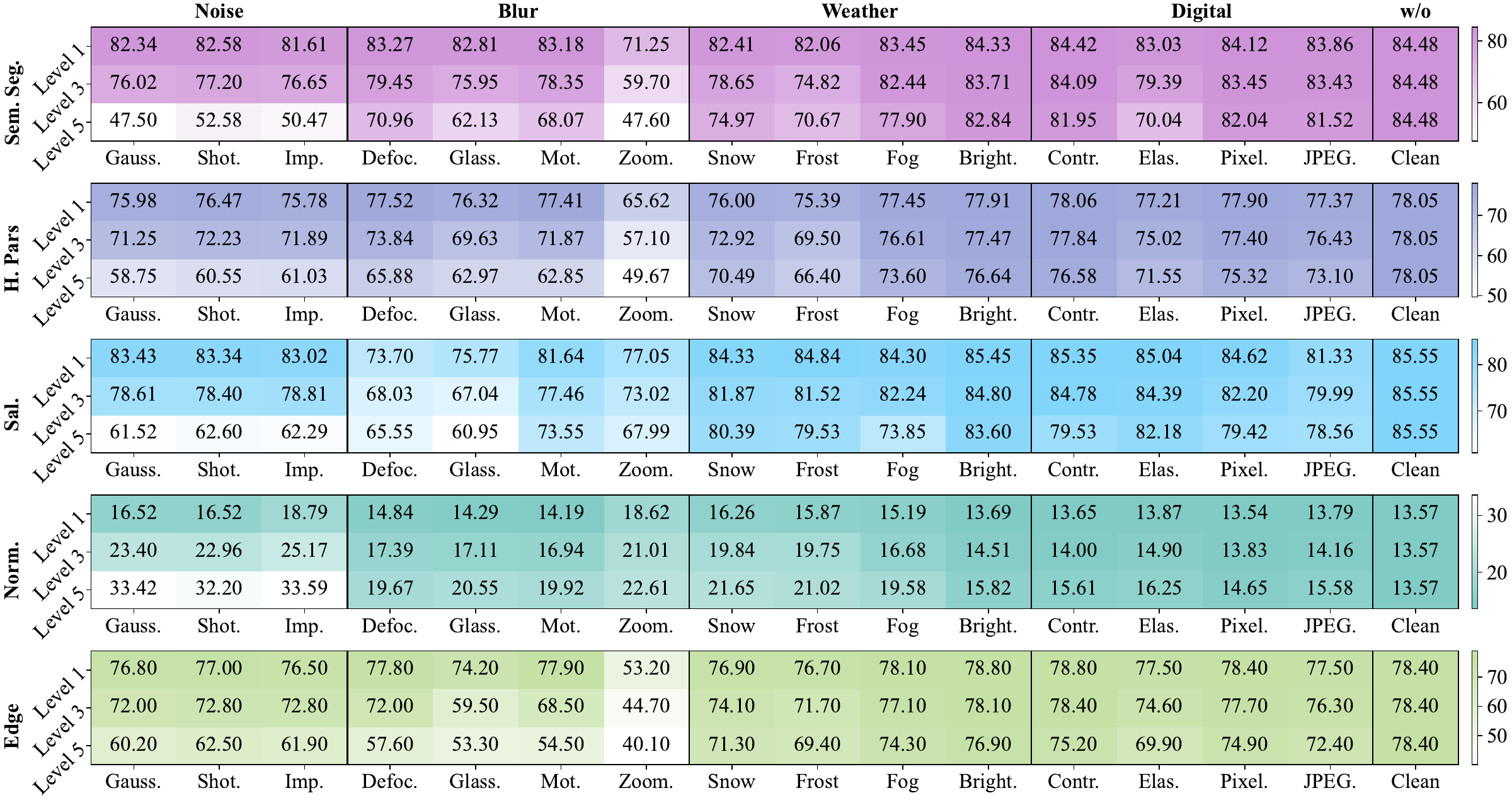}
    \caption{
    \textbf{Corruption Robustness Evaluation on Pascal-Context.}
Across five tasks, performance decreases with corruption severity, with zoom blur and gaussian noise inducing the largest degradations (e.g., Sem. Seg. drops from 84.48 on clean to 47.50 at Level 5). Weather/digital corruptions are comparatively mild at Levels 1–3, while normal error rises primarily under noise.}
% Our method maintains stable performance across multiple tasks under diverse corruption types and severity levels (Level 1, 3, and 5), demonstrating strong robustness to noise, blur, weather, and digital degradations.}
    \label{fig:corr_heatmaps}
\end{figure*}

\begin{figure*}
    \centering
    \includegraphics[width=0.795\linewidth]{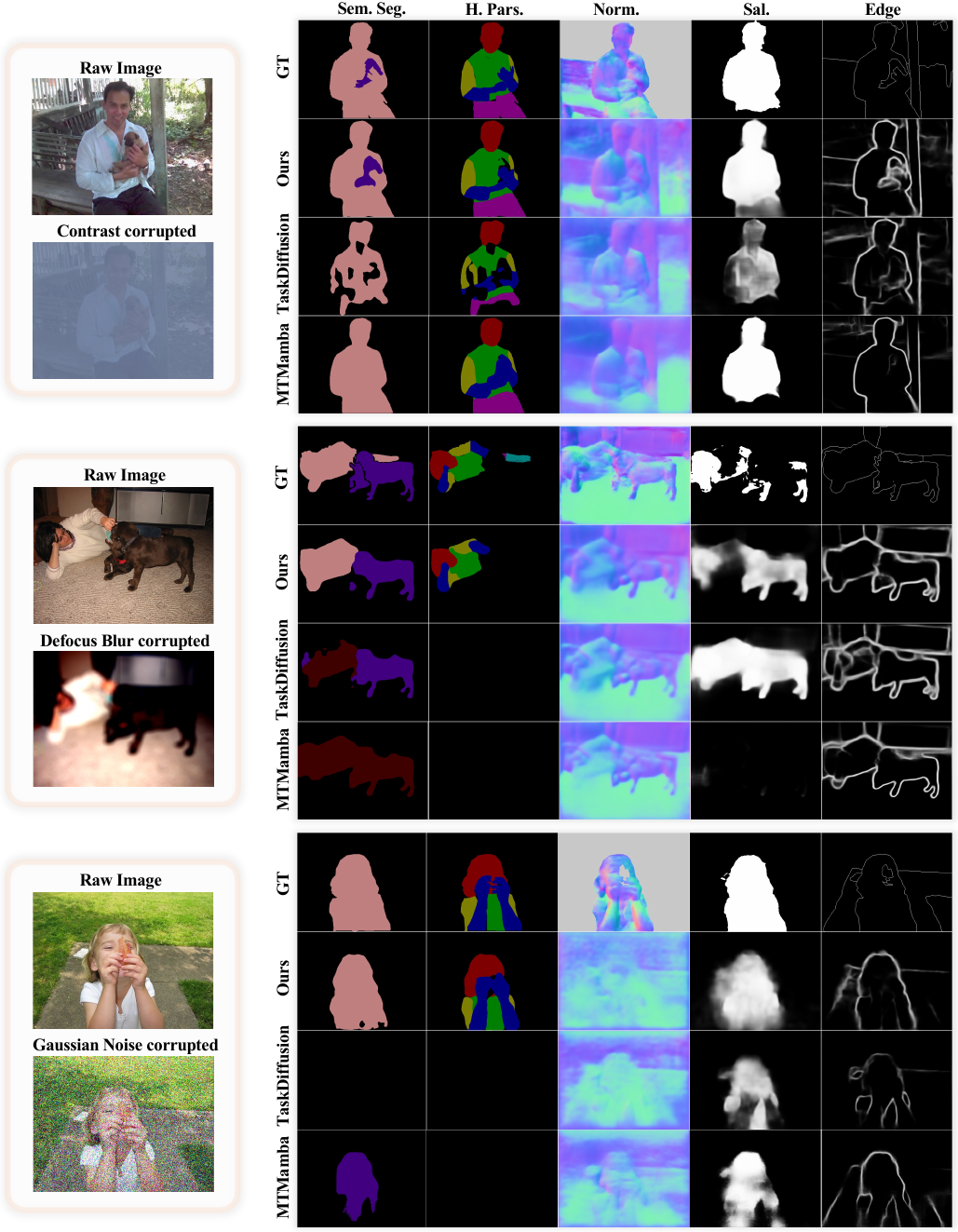}
    \caption{\textbf{Corruption Robustness Comparison with TaskDiffusion~\cite{yangmulti}.} All examples are drawn from the Pascal-Context dataset under level-5 corruptions. Our method demonstrates stronger robustness to severe degradations, yielding cleaner object boundaries and more stable predictions.}
    \label{fig:corruption_vis}
\end{figure*}

\subsubsection{Hyperparameter Analysis}
\label{sec4.4.2}
We quantify how hyperparameters shape the capacity--efficiency trade-off in Fig.~\ref{fig:hyperparameters}, varying TG-LoRA rank, TA-MoE expert count, TA-MoE LoRA rank, and Top-$k$ routing.

\textbf{LoRA Rank of TG-LoRA.}
Increasing TG-LoRA rank from $2$ to $16$ (Fig.~\ref{fig:hyperparameters}(a)) improves all tasks (Sem. Seg. $83.40 \rightarrow 84.48$ mIoU, H. Pars. $77.66 \rightarrow 78.05$ mIoU, Sal. $84.73 \rightarrow 85.55$ maxF). Consistently, $\Delta_m$ increases from $8.07\%$ to $8.88\%$, indicating stronger overall multi-task performance. All parameters rise modestly ($355.5$M $\rightarrow$ $357.6$M). At rank $32$, gains saturate or reverse (e.g., Sem. Seg. $84.35$) despite higher parameters/FLOPs, indicating over-parameterization without improved alignment.

\textbf{Number of Experts in TA-MoE.}
{
\looseness=-1
Varying experts (Fig.~\ref{fig:hyperparameters}(b)) shows a similar saturation: $2$$\rightarrow$$16$ experts improves segmentation ($83.47 \rightarrow 84.48$ mIoU) and boundary detection ($78.30 \rightarrow 78.50$ odsF) with limited parameter growth ($355.9$M $\rightarrow$ $357.6$M). The aggregated gain peaks at $16$ experts with $\Delta_m=8.88\%$, suggesting the best balance between specialization capacity and efficiency. At $32$ experts, performance saturates or slightly decreases (e.g., parsing $78.05 \rightarrow 77.64$), suggesting redundancy rather than stronger specialization.
\par
}

\begin{table}
\centering
\caption{Comparison of backbone variants under the same multi-task setting.}
% Ablation study of different backbone variants under the same multi-task setting.}
\vspace{-2mm}
\label{Backbone Ablation}
\scalebox{0.9}{
\begin{tabular}{ccccccc}
\toprule
 \multirow{2}{*}{\textbf{Backbone}} & \textbf{Para.} & \textbf{Sem. Seg.} & \textbf{Pars.} & \textbf{Sal.} & \textbf{Norm.} & \textbf{Edge} \\%& $\Delta_{m}[\%]$ \\
  & (M) & mIoU$\uparrow$ & mIoU$\uparrow$ & maxF$\uparrow$ & mErr$\downarrow$ & odsF$\uparrow$ \\%& $\uparrow$ \\
\cmidrule(lr){1-1}\cmidrule(lr){2-2}\cmidrule(lr){3-7}
% MTL Baseline &  & 81.26 & 72.48 & 84.66 & 15.60 & 76.70 & 0.00 \\
% \cdashline{1-8}
ViT-S  & 62.0 & 80.62 & 73.36 & \underline{85.28} & 13.79  & 76.40 \\%& 2.47 \\
ViT-B  & 126.6 & \textbf{85.25} & 75.66 & 84.95 & 13.66  & 77.50 \\%& 4.62 \\
ViT-L  & 357.6 & 84.48 & \underline{78.05} & \textbf{85.55} & \textbf{13.57}  & \underline{78.40} \\%& \textbf{5.59}  \\
ViT-H  & 911.3 & \underline{84.80} & \textbf{78.87} & 83.17 & \underline{13.62}  & \textbf{79.60} \\%& \underline{5.58} \\
 \hline
\end{tabular}
}
\end{table}
\begin{table*}[!t]
\centering
% \Huge
\caption{Comparison with MLoRE using a DINOv3 backbone.}
\vspace{-2mm}
\label{compare_mlore}
% \scalebox{0.5}{
\resizebox{0.8\textwidth}{!}{

\begin{tabular}{cccccccccccc}
\toprule
 \multirow{2}{*}{\textbf{Modules}} & \textbf{Sem. Seg.} & \textbf{H. Pars.} & \textbf{Sal.} & \textbf{Norm.} & \textbf{Edge} & $\Delta_{m}$ & \textbf{Para.} & \textbf{Train. Para.} & \textbf{Flops} & \textbf{Latency} \\
 & mIoU$\uparrow$ & mIoU$\uparrow$ & maxF$\uparrow$ & mErr$\downarrow$ & odsF$\uparrow$ & $(\textit{\%})\uparrow$ & $(\textit{M})\downarrow$ & $(\textit{M})\downarrow$ &(\textit{G})$\downarrow$ & (\textit{ms})$\downarrow$ \\
\cmidrule(lr){1-1}\cmidrule(lr){2-6}\cmidrule(lr){7-7}\cmidrule(lr){8-11}
MTL Baseline & 82.61 & 76.38 & 77.09 & 17.63 & 74.00 & 0.00 & \textbf{328} & \textbf{25} & \textbf{759.7} & \textbf{198.9}  \\
% \cdashline{1-7}
MLoRE~\cite{yang2024multi} & \underline{83.85} & \textbf{78.05} & \underline{84.98} & \underline{15.51} & \underline{78.10} & \underline{6.30} & 407 & 407 & \underline{829.9} & \underline{203.2} \\
\ours & \textbf{84.48} & \textbf{78.05} & \textbf{85.55} & \textbf{13.57} & \textbf{78.40} & \textbf{8.88} & \underline{358} & \underline{54} & 1029.9 & 224.2 \\
 \bottomrule
\end{tabular}}
% }
\end{table*}
\begin{table*}
\centering
\small
\caption{Comparison with shared LoRA, task-specific LoRA and decoder-side MoE under matched trainable-parameter budgets.}
\label{Table:Ablation}
\scalebox{0.8}{
\begin{tabular}{ccccccccccc}
\toprule
\multirow{2}{*}{\textbf{Method}} & 
\textbf{Sem. Seg.} & 
\textbf{H. Pars.} & 
\textbf{Sal.} & 
\textbf{Norm.} & 
\textbf{Edge} & 
$\Delta_{m}$ & 
\textbf{Backbone} & 
\textbf{Decoder} & 
\textbf{Train.} & 
\textbf{Para.} \\

\rowcolor{revisionyellow} & mIoU$\uparrow$ 
& mIoU$\uparrow$ 
& maxF$\uparrow$ 
& mErr$\downarrow$ 
& odsF$\uparrow$ 
& $(\%)\uparrow$ 
& \textbf{Train.} (M)$\downarrow$
& \textbf{Train.} (M)$\downarrow$
& \textbf{Para.} (M)$\downarrow$
& (M)$\downarrow$ \\

\cmidrule(lr){1-1}
\cmidrule(lr){2-6}
\cmidrule(lr){7-7}
\cmidrule(lr){8-11}

MTL Baseline
& 82.61
& 76.38
& 77.09
& 17.63
& 74.00
& 0.00
& 25.2
& 0
& 25.2
& 328.4 \\

\cmidrule(lr){1-1}
\cmidrule(lr){2-6}
\cmidrule(lr){7-7}
\cmidrule(lr){8-11}

% A: Shared LoRA ($r=171$)
Shared LoRA
& 83.60
& 77.52
& \underline{85.19}
& \textbf{13.51}
& \underline{78.30}
& \underline{8.48}
& 25.2
& 29.2
& 54.4
& 357.6 \\

% B: Task LoRA ($r=186$)
Task LoRA
& \underline{84.25}
& 77.95
& 84.65
& 13.77
& \underline{78.30}
& 8.31
& 25.2
& 29.2
& 54.4
& 357.6 \\

% C: Decoder MoE ($r=111$)
Decoder MoE
& 83.22
& \underline{78.00}
& 82.46
& 15.16
& 78.20
& 5.90
& 0
& 54.4
& 54.4
& 357.6 \\

\cmidrule(lr){1-1}
\cmidrule(lr){2-6}
\cmidrule(lr){7-7}
\cmidrule(lr){8-11}

\ours 
& \textbf{84.88} 
& \textbf{78.05} 
& \textbf{85.55} 
& \underline{13.57} 
& \textbf{78.40} 
& \textbf{8.88}
& 25.2 
& 29.2 
& 54.4 
& 357.6 \\

\bottomrule
\end{tabular}
}
\end{table*}

\textbf{LoRA Rank of TA-MoE.}
Fig.~\ref{fig:hyperparameters}(c) indicates that moderate per-expert rank is sufficient: rank $4$ achieves strong overall accuracy (segmentation $84.48$ mIoU) and the lowest normal error ($13.57$ mErr) at $357.6$M parameters. It also obtains the highest aggregated gain ($\Delta_m=8.88\%$), confirming that moderate rank is sufficient for overall multi-task capacity. Lower ranks underfit task-specific refinement, whereas higher ranks increase parameters/FLOPs with marginal, unstable gains.

\textbf{Top-$k$ Value in TA-MoE.}
Fig.~\ref{fig:hyperparameters}(d) shows that increasing $k$ from $2$ to $8$ improves performance, peaking at $84.48$ mIoU (segmentation) and $85.55$ maxF (saliency) with unchanged parameters. $\Delta_m$ also reaches the best value of $8.88\%$ when $k=8$. Larger $k$ (e.g., $12$ or $16$) yields marginal gains or mild degradation, consistent with reduced sparsity-driven specialization under broader expert activation.

Collectively, these ablations identify a coherent regime where improvements arise from structured task-adaptive design, including TG-LoRA alignment, sparse TA-MoE specialization, and SPI prior injection, rather than from indiscriminate increases in trainable capacity.

\begin{table}
\centering
\caption{Averaged robustness comparison on PASCAL-Context across all corruptions.}
\label{tab:overall_robustness}
{
\scalebox{0.9}{
\begin{tabular}{cccccc}
\toprule
\multirow{2}{*}{\textbf{Method}} & 
\textbf{Sem. Seg.} & \textbf{H. Pars.} & \textbf{Sal.} & \textbf{Norm.} & \textbf{Edge} \\
& mIoU$\uparrow$ & mIoU$\uparrow$ & maxF$\uparrow$ & mErr$\downarrow$ & odsF$\uparrow$ \\
\midrule
MTL Baseline & \underline{65.36} & \underline{64.54} & \underline{70.08} & \underline{22.45} & \underline{63.07} \\
TaskDiffusion & 57.01 & 49.48 & 67.42 & 22.81 & 59.04 \\
\ours & \textbf{68.08} & \textbf{67.03} & \textbf{72.77} & \textbf{21.47} & \textbf{64.96} \\

\bottomrule
\end{tabular}
}}
\end{table}

\begin{figure}
    \centering
    \includegraphics[width=\linewidth]{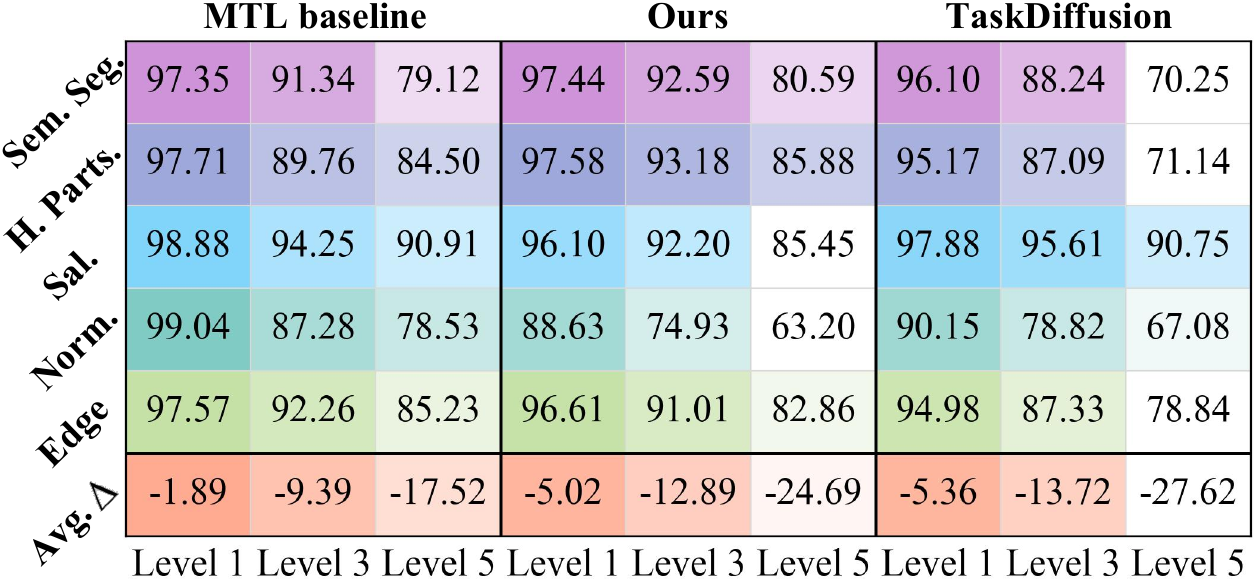}
    \caption{\textbf{Clean-normalized Performance Retention under Corruption.}
Task-wise retention for the MTL baseline, D$^2$MoE, and TaskDiffusion. Avg.~$\Delta$ denotes mean relative degradation from clean performance.}
    \label{fig:retention}
\end{figure}

\subsection{Feature Visualization and Analysis}
\label{sec4.5}
\textbf{Feature Embedding Visualization (t-SNE).}
Fig.~\ref{fig:tsne_semseg} shows 3D t-SNE projections of the learned feature embeddings, directly comparing the MTL baseline (top row) with \ours (bottom row). \ours reshapes the embedding manifold into task-aligned submanifolds rather than an entangled cloud. On Sem. Seg. (Pascal-Context, NYUD-v2, Cityscapes), \ours forms tighter, well-separated clusters with higher intra-class compactness and reduced cross-class overlap than the MTL baseline, reflecting increased inter-class variance and less negative transfer. This behavior matches selective routing within a shared backbone: early layers preserve broadly reusable cues (edges, textures), while routed experts enhance class-conditional directions, widening margins between semantic modes while maintaining the global manifold structure. Regarding depth estimation, \ours yields clearer and more structured groupings, suggesting routing induces a more geometry-aware embedding where features are organized topologically according to their respective depth bins, a marked improvement over the disordered feature clouds observed in the baseline.

\noindent\textbf{Router Correlation Analysis.}
Fig.~\ref{fig:router_vis} reveals stage- and operator-specific specialization. In QKV projections, early stages display strong cross-task coupling (off-diagonal correlations), whereas deeper stages become near diagonal-dominant, implying \ours progressively privatizes attention access patterns (queries/keys/values) to mitigate negative transfer in information selection. Conversely, attention output projections exhibit structured, non-monotonic sharing: an early block structure (notably among Hum./Nor./Sem.) persists, and late-stage correlations remain strong for specific pairs (e.g., Nor.--Sem., Hum.--Sal.), indicating \ours avoids forced full separation and reuses experts when supervision is complementary. Overall, TA-MoE specializes where tasks attend (QKV) while selectively sharing what is read out (output projection), disentangling conflicts while enabling transfer when tasks align.

\subsection{Robustness to Distribution Shift}
\label{sec4.6}
We assess domain generalization on Pascal-Context under the corruption benchmark of ImageNet~\cite{hendrycks2019benchmarking}. Fig.~\ref{fig:corr_heatmaps} reports corruption-wise performance across severities ($1$/$3$/$5$), and Fig.~\ref{fig:corruption_vis} visualizes level-$5$ out-of-distribution (OOD) predictions against TaskDiffusion~\cite{yangmulti} and MTMamba~\cite{lin2024mtmamba}. To separate corruption sensitivity from gains due to higher clean accuracy, Fig.~\ref{fig:retention} additionally reports clean-normalized performance retention, while Tab.~\ref{tab:overall_robustness} reports severity-5 results averaged over all 15 corruptions.
% Higher values indicate smaller degradation from clean performance

% To avoid selection bias, we further report quantitative comparisons with the MTL baseline and TaskDiffusion in Tab.~\ref{tab:overall_robustness}, where results are averaged over all 15 corruptions at severity level 5. \ours achieves the best performance across all five tasks.

Across all tasks, the dominant trend in Fig.~\ref{fig:corr_heatmaps} is graded performance change with increasing corruption severity, indicating stable behavior under covariate shift. Noise corruptions are consistently the most damaging: Sem. Seg. drops from $82.34$ (Gaussian, level-$1$) to $47.50$ (level-$5$), and H. Pars. from $75.98$ to $58.75$ under the same shift, while surface-normal error increases smoothly from $16.52$ (clean) to $33.42$ (Gaussian, level-$5$). In the blur family, zoom blur is the primary failure case across tasks (e.g., Sem. Seg. $71.25 \rightarrow 47.60$, H. Pars. $65.62 \rightarrow 49.67$, Edge $53.20 \rightarrow 40.10$ from level-1 to level-$5$), whereas defocus/motion blur produce substantially smaller degradations. Weather and digital corruptions are comparatively benign for most tasks: for Sem. Seg., level-$5$ contrast remains close to clean ($81.95$ vs.\ $84.48$), and saliency stays high across many level-$5$ weather/digital shifts, demonstrating broad OOD robustness beyond a single corruption type.

Tab.~\ref{tab:overall_robustness} shows that D$^2$MoE achieves the best average corrupted performance across all five tasks, outperforming both the MTL baseline and TaskDiffusion.
Fig.~\ref{fig:retention} provides a complementary normalized comparison: D$^2$MoE shows smaller average degradation than TaskDiffusion at all severities ($-5.02\%/-12.89\%/-24.69\%$ vs. $-5.36\%/-13.72\%/-27.62\%$), with stronger retention particularly on Sem. Seg. and H. Pars., whereas the MTL baseline exhibits smaller overall degradation ($-1.89\%/-9.39\%/-17.52\%$). Thus, D$^2$MoE's superior corrupted performance stems from both stronger clean performance and improved retention over TaskDiffusion, rather than uniformly lower corruption sensitivity.
% higher corrupted accuracy reflects both its stronger clean performance and improved retention on selected tasks, rather than uniformly lower corruption sensitivity.}

Fig.~\ref{fig:corruption_vis} qualitatively confirms these trends under severe shift. Under contrast change and defocus blur, \ours preserves coherent region topology in Sem. Seg. and H. Pars. and maintains continuous edge structure. Even under Gaussian noise which corresponds to the largest metric drops, \ours retains structured masks, plausible part layouts, and recognizable boundaries, whereas TaskDiffusion~\cite{yangmulti} exhibits region fragmentation, unstable part layouts, and degraded boundary continuity. Compared with MTMamba~\cite{lin2024mtmamba}, \ours better preserves foreground completeness and task-consistent structures, avoiding over-smoothing and missing responses, especially in saliency and edge predictions.

Overall, the absolute and clean-normalized results provide complementary evidence: D$^2$MoE maintains stronger absolute OOD performance and degrades less than TaskDiffusion on average, while the MTL baseline exhibits smaller overall normalized degradation despite D$^2$MoE showing better retention on selected tasks.

\section{Conclusion}
\label{sec:5}
In this work, we addressed the challenge of multi-task dense prediction in the vision foundation model paradigm, where heterogeneous tasks must be adapted without compromising the robustness of pretrained representations. We proposed \ours, which introduces a structured Mixture of Task Adapters to decouple task-generic alignment from task-specific specialization atop a frozen DINOv3. By combining Task-Generic Low-Rank Adaptation across all attention layers with a Task-Adaptive Mixture of Experts selectively applied in later stages, \ours resolves task conflicts while preserving a shared semantic prior within a single forward pass. In addition, semantic prior injection further improves structural coherence and cross-task consistency. Extensive experiments across diverse benchmarks demonstrate that this design effectively mitigates negative transfer and achieves strong multi-task performance with high parameter efficiency. Overall, our results highlight the importance of layer-aware, conditional adaptation for unlocking the full potential of vision foundation models in multi-task dense prediction.

% \section{Where to get \LaTeX \ help - user groups}
% \noindent The following on-line groups are very helpful to beginning and experienced \LaTeX\ users. A search through their archives can provide many answers to common questions.
% \begin{list}{}{}
% \item{\url{http://www.latex-community.org/}} 
% \item{\url{https://tex.stackexchange.com/} }
% \end{list}

%%%
\bibliographystyle{IEEEtran}
\bibliography{ieeetran}

% \section{Biography Section}

\end{document}